\documentclass[pdflatex,sn-mathphys-num]{sn-jnl}

\usepackage{graphicx}%
\usepackage{multirow}%
\usepackage{amsmath,amssymb,amsfonts}%
\usepackage{amsthm}%
\usepackage{mathrsfs}%
\usepackage[title]{appendix}%
\usepackage{textcomp}%
\usepackage{manyfoot}%
\usepackage{booktabs}%
\usepackage{algorithm}%
\usepackage{algorithmicx}%
\usepackage{float}
\usepackage{tabularx}
\usepackage{algpseudocode}%
\usepackage{listings}%

\usepackage{adjustbox}
\usepackage{booktabs}

\usepackage{listings}
\theoremstyle{thmstyleone}%
\theoremstyle{thmstyletwo}%

\theoremstyle{thmstylethree}%

\begin{document}

\title[Article Title]{Evalita-LLM: Benchmarking Large Language Models on Italian}


\author*[1]{\fnm{Bernardo} \sur{Magnini}}\email{magnini@fbk.eu}

\author[1]{\fnm{Roberto} \sur{Zanoli}}\email{zanoli@fbk.eu}
\equalcont{These authors contributed equally to this work.}

\author[2]{\fnm{Michele} \sur{Resta}}\email{michele.resta@igenius.ai}
\equalcont{These authors contributed equally to this work.}

\author[2]{\fnm{Martin} \sur{Cimmino}}\email{martin.cimmino@igenius.ai}
\equalcont{These authors contributed equally to this work.}

\author[2]{\fnm{Paolo} \sur{Albano}}\email{paolo.albano@igenius.ai}
\equalcont{These authors contributed equally to this work.}

\author[3]{\fnm{Marco} \sur{Madeddu}}\email{marco.madeddu@unito.it}
\equalcont{These authors contributed equally to this work.}

\author[3]{\fnm{Viviana} \sur{Patti}}\email{viviana.patti@unito.it}
\equalcont{These authors contributed equally to this work.}

\affil*[1]{\orgdiv{Fondazione Bruno Kessler},  \orgaddress{\street{Via Sommarive 18},  \state{Povo, Trento}, \country{Italy}}}

\affil[2]{\orgdiv{iGenius}, \orgaddress{\street{Via Principe Amedeo 5}, \state{Milan}, \country{Italy}}}

\affil[3]{\orgdiv{University of Torino}, \orgname{Computer Science Dept.}, \orgaddress{\street{Corso Svizzera 185},  \state{Torino}, \country{Italy}}}


\abstract{We describe Evalita-LLM, a new benchmark designed to evaluate Large Language Models (LLMs) on Italian tasks. The distinguishing and innovative features of Evalita-LLM are the following: (i) all tasks are native Italian, avoiding issues of translating from Italian and potential cultural biases; (ii) in addition to well established multiple-choice tasks, the benchmark includes generative tasks, enabling more natural interaction with LLMs; (iii) all tasks are evaluated against multiple prompts, this way mitigating the model sensitivity to specific prompts and allowing a fairer and objective evaluation. We propose an iterative methodology, where candidate tasks and candidate prompts are validated against a set of LLMs used for development. We report experimental results from the benchmark's development phase, and provide performance statistics for several state-of-the-art LLMs.}




\maketitle

\section{Introduction}\label{sec:intro}

Recently, several Large Language Models (LLMs) have been made available that are trained on Italian data, and several others have been announced for the next months. Those models (e.g., LLaMAntino \cite{polignano2024advanced}, the Minerva family\footnote{\url{https://nlp.uniroma1.it/minerva/blog}}, Italia\footnote{\url{https://huggingface.co/iGeniusAI/Italia-9B-Instruct-v0.1}}) are released as open source, although with different licenses, and are all available through the Hugging Face (HF) platform\footnote{\url{https://huggingface.co}}. In this context, performance evaluation of such models through appropriate benchmarking is becoming crucial. However, in the current evaluation practice, LLMs for Italian are mostly evaluated on benchmarks automatically translated from English, which poses critical issues:

 \begin{itemize}
     \item \textit{Quality of translations.} Translation from English is carried out using available automatic translators, whose quality is rarely checked by professionals. In \cite{Moroni2024}, several issues related to the translation of English benchmarks are highlighted. Furthermore, the code and models used to translate these benchmarks are not directly available, making it hard – if not impossible – to reproduce the translations. This, in turn, makes it difficult to analyze whether there are errors or if there is a margin for improvement in the translation process originally used to translate the benchmarks.
    \item \textit{Content}. Although translated into Italian, benchmarks are still based on anglo-american texts, which do not consider the crucial diversity of Italian culture. As an example, in the MMLU benchmark \cite{hendryckstest2021}, several questions refer to very specific facts about the anglo-american culture, which are hardly known by LLMs trained only on Italian data. The presence of such cultural biases can potentially alter the evaluation process.
    \item \textit{Style}. Translated benchmarks are very sensitive to the style of automatic translation and do not fully reflect the stylistic varieties of native Italian. Particularly for generative tasks (e.g., producing a summary), stylistic biases may affect the evaluation results.
 \end{itemize}

At the time of writing, the landscape of evaluation is rapidly evolving. As an example, the Ita-bench\footnote{\url{https://huggingface.co/collections/sapienzanlp/ita-bench-italian-benchmarks-for-llms-66337ca59e6df7d7d4933896}} is now available on Hugging Face, including tasks originally developed in Evalita and adapted to LLMs. In addition, the Calamita initiative \cite{attanasio2024calamita} has produced several benchmarks for Italian LLMs, opening new options for a more comprehensive evaluation. In this context, our contribution is to provide a methodologically well-founded benchmark, exploiting as much as possible of the potential behind generative LLMs for Italian.
Evalita-LLM is freely available on HF and, in our intention, will be one of the reference benchmarks for LLMs as far as the Italian language is concerned.

\paragraph{Issues addressed in the paper}
In addition to exploring the behavior of recent LLMs for Italian, the paper addresses some relevant methodological issues related to benchmarking LLMs, which can be applied to languages other than Italian.

\begin{itemize}
    \item \textit{Generative tasks.} Prompting for text generation is the more natural way (i.e., the most similar to human behavior) to interact with LLMs. However, the current practice in benchmarking LLMs is still oriented towards casting tasks as multiple-choice selection, where the best option is the one with the highest probability, usually estimated through the logits of the model for a particular choice. While this method is simple and provides a clear scoring mechanism, it can not be applied to all tasks (e.g., a good summary is not the most probable sequence of sentences), and in many cases produces artificial formulations of a task (e.g., when Named Entity Recognition is implemented as a choice of entity types for each token in a sentence). On the other hand, generative tasks suffer from two interrelated issues: first, the current metrics (e.g., BLEU, ROUGE, BERTScore, COMET) are still a poor approximation of human capacities to extrapolate the relevant information from a complex output; second, it is still hard to prompt the model to output its results in a precise format (e.g., attribute-value pairs), which can be parsed to extract the result of a task execution. The two issues together, low quality metrics and poor output format, make it challenging to implement as many generative tasks as desired. 
    \item \textit{Multiple-prompts.} Both multiple-choice and generative tasks are typically implemented using a single prompt. However, a single prompt may work well with a model and very badly with another model, raising the question of the fairness of single-prompt evaluation. In addition, there are no established guidelines on how a single prompt is designed for different tasks: for instance, prompts are often just copy-and-pasted from different datasets, without any attempt to understand how different prompts work with respect to a certain task.
    \item \textit{Dataset saturation.} Determining the degree of contamination of the data used both to test a model and for pre-training that model, is highly challenging. Furthermore, it is often unclear whether a certain prompt has been used during the instruction phase of the model. If the prompt is known by the model, we may expect better performance, which affects the evaluation. Even more critical, the relationship between data saturation and prompt saturation has not yet been sufficiently investigated.
\end{itemize}

\section{Methodology}
\label{sec:methodology}

In order to avoid the effort to produce new annotated data, Evalita-LLM takes advantage of existing datasets developed in the last fifteen years under Evalita\footnote{\url{https://www.evalita.it}}, an initiative of the Italian Association for Computational Linguistics (AILC\footnote{\url{https://www.ai-lc.it}}). Thanks to the contributions of the Italian community on Computational Linguistics, Evalita has produced about 70 datasets covering various linguistic phenomena, out of which about 35 are available under open licenses through the European Language Grid platform (ELG)\footnote{\url{https://live.european-language-grid.eu/}}, thanks to the Evalita4ELG initiative run by the University of Turin. 

Evalita-LLM is implemented using the lm-evaluation-harness library\footnote{\url{https://github.com/EleutherAI/lm-evaluation-harness}} \cite{biderman2024lessonstrenchesreproducibleevaluation} available in HF. We followed three steps to develop a task using lm-evaluation-harness:

\begin{itemize}
    \item \textit{Dataset selection}: a dataset must first be identified on the ELG catalogur, or other sources, then converted into the required HF format and finally uploaded to HF.
    \item \textit{Task definition}: once a dataset is uploaded on HF, a task has to be defined, which includes several aspects: (i) linking to the dataset; (ii) pre-processing the data; (iii) designing the LLM prompt: (iv) zero-shot or few-shot for in-context learning; (v) implementing a post-processing script; (vi) defining the evaluation metrics.
    Tasks are defined either as zero-shot or few-shot only. Under the assumption that most of the LLMs to be evaluated will be instruction-tuned, Evalita-LLM does not include tasks with training data for fine-tuning.
    \item \textit{Model evaluation}: while the final goal is to evaluate LLMs against Evalita-LLM, we foresee an incremental process to validate an effective set of tasks. In this process, a candidate task is tested (e.g., to validate prompt efficiency) on available models for Italian during the development phase.
\end{itemize}

\begin{figure}
    \centering
    \includegraphics[width=0.8\linewidth]{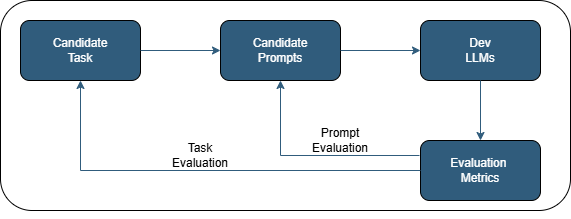}
    \caption{Evalita-LLM incremental validation methodology.}
    \label{fig:methodology}
\end{figure}

Figure \ref{fig:methodology} shows the main steps of the Evalita-LLM development. The starting point is a set of candidate datasets already available through the Evalita campaigns and other initiatives. We consider only Italian native datasets, which are the large majority. We started with about 15 candidate datasets, half of them naturally fitting with the multiple-choice setting and the other half fitting with the generative setting. Given a candidate task, the second step is to design a set of candidate prompts for the task. Initially, prompts were derived from similar tasks already implemented, e.g., for English. To test the effectiveness of the candidate prompts, we used a set of LLMs with similar characteristics (see details in Section \ref{sec:dev-llms}) and few evaluation metrics (see Section \ref{sec:metrics}). The process is incremental in the sense that both the initial candidate tasks and their associated prompts can be incrementally refined according to the performance obtained on the dev LLMs. For instance, if a prompt is not well understood by any of the assessed LLMs, then that prompt is discarded and a different variant is tested. Similarly, a task that reveals itself too complex for the dev LLMs is not selected to be included in the benchmark. A candidate task and an associated set of prompts are selected when some minimum criteria (see discussion in Section \ref{sec:experiments}) are achieved, including a reasonable execution time on standard hardware.

At the end of the process we selected 10 tasks for Evalita-LLM, representing different linguistic phenomena, linguistic genres and different applications, which are presented in the next Section.

\section{Evalita-LLM Tasks}\label{sec:tasks}

The following 10 tasks, reported in Table \ref{tab:tasks} are available for the first Evalita-LLM release. Most of the tasks have been proposed at the Evalita initiative, while two of them (admission tests and summarization) were not. Out of ten tasks, six have been implemented as multiple-choice, where the model is asked to select one option from a set of possible choices, and four are implemented through the \textit{generate-until} method, where the model is asked to generate the task solution. Tasks have been selected to be representative of different core linguistic competences, including word sense disambiguation, text classification, question answering, information extraction and summarization. They are also representative of textual documents in different domains, including news, social media and wiki, scientific documents and public administration. Metrics include accuracy, $F_1$ and rouge for summarization.

Although the data are generally of good quality, a number of adaptations were required to use them to evaluate LLMs and to implement them in the lm-evaluation-harness library\footnote{\url{https://github.com/EleutherAI/lm-evaluation-harness}} \cite{biderman2024lessonstrenchesreproducibleevaluation}.
For each task, we report a brief description, the information about the original corpus, the performance of baseline models, the processing steps we applied to obtain a suitable version for the evaluation in the library and the details of the processed dataset. 

For the generative tasks, we also report the characters used as stop tokens. This is important because LLMs may give a correct answer and then continue generating text that is not inherent to the task, which can cause a decrease in performance.

\begin{table}[ht]
\centering
\begin{tabular}{c|l|l|l|l|l}
\toprule
\textbf{} & \textbf{Task}  & \textbf{Core Competence} & \textbf{Domain} & \textbf{LLM eval} & \textbf{Metric} \\ \midrule
1 & Word in context   & Word  disambiguation & news & multiple-choice & $F_{1}$ \\ 
2 & Textual entailment  & Semantic inference & news & multiple-choice & accuracy \\ 
3 & Sentiment analysis & Text classification  & social & multiple-choice & $F_1$-macro \\ 
4 & Hate speech  & Text classification & social & multiple-choice & $F_1$-macro \\ 
5 & FAQ & Question answering   & P.A. & multiple-choice & accuracy \\ 
6 & Admission tests  & Question answering & scientific & multiple-choice & accuracy  \\ 
7 & Lexical substitution & Word disambiguation & news & generate-until & $F_{1}$  \\ 
8 & Entity recognition & Information extraction & mixed & generate-until & $F_{1}$ \\ 
9 & Relation extraction  & Information extraction & scientific & generate-until & $F_{1}$ \\
10 & Summarization  & Text generation & wiki & generate-until & rouge \\ \botrule
\end{tabular}
\caption{Tasks in the Evalita-LLM benchmark.}
\label{tab:tasks}
\end{table}

\subsection{Word in Context (WIC)}
The Word in Context (WiC) task, proposed at Evalita 2023\footnote{\url{https://wic-ita.github.io/task}}, focuses on word sense disambiguation in context. A comprehensive description of the task can be found in~\cite{Cassotti2023}. This task is composed of two sub-tasks: binary classification (Sub-task 1) and ranking (Sub-task 2). This paper focuses on Sub-task 1, a binary classification task to determine if a word \textit{w} in two sentences, \textit{s1} and \textit{s2}, has the same meaning.

\paragraph{Examples}
Some examples for the Word In Context task are reported in Table \ref{tab:wic_examples}.

\setlength\tabcolsep{5pt}
\begin{table*}[h!]
\centering
    \begin{tabularx}{\textwidth}{XX}
    \toprule
    \textbf{Italian} & \textbf{Translated in English} \\
    \midrule
    \multicolumn{2}{>{\hsize=\dimexpr2\hsize+2\tabcolsep+\arrayrulewidth\relax}c}{\textbf{Same meaning}}\\
    \\
     \textbf{s1:}Basta con lo stillicidio \textit{pilotato} delle notizie & \textbf{s1:}Enough with the \textit{controlled} drip-feed of news \\
    \textbf{s2:} Scandalo, indagato il consigliere comunale. Tre arresti per appalti \textit{pilotati} e corruzione.  & \textbf{s2:} Scandal, city councilor investigated. Three arrested for \textit{rigged} tenders and corruption. \\
    \midrule

    \multicolumn{2}{>{\hsize=\dimexpr2\hsize+2\tabcolsep+\arrayrulewidth\relax}c}{\textbf{Different Meaning}}\\
    \\
    \textbf{s1:} Una successione di strade romane identica a quelle indicate in un file del \textit{palmare} sequestrato. & \textbf{s1:} A succession of Roman roads identical to those indicated in a file from the seized \textit{PDA}. \\
    \textbf{s2:} A meno che non si tratti di farli di una evidenza \textit{palmare}, gli estensori fanno bene a non aver fretta nella redazione del verbale stesso. & \textbf{s2:} Unless it is a matter of making them a matter of \textit{clear} evidence, the writers are right not to rush into drafting the minutes themselves. \\
    
    \bottomrule
    \end{tabularx}
\caption{Examples from the Word in Context dataset.}
\label{tab:wic_examples}
\end{table*}




\paragraph{Original dataset}
For model development, the following datasets are available\footnote{\url{https://github.com/wic-ita/data}}: a training dataset containing 2,805 examples used to train the model; a development dataset comprising 500 examples, suitable for model evaluation during the training phase, such as hyperparameter tuning; a monolingual test dataset with 500 examples; and a cross-lingual test dataset, also with 500 examples. This study exclusively focuses on the monolingual dataset.

\paragraph{Evaluation Measure}
Sub-task 1 is evaluated using the $F_1$-macro score, where the system's binary classification predictions are compared against the ground truth. The system predicts ‘‘1’’ when the word has the same meaning across the two sentences and ‘‘0’’ when it does not. $F_1$-macro is computed for each class, and the average of these scores is taken to obtain the $F_1$-macro score.


\paragraph{Results} The baseline model employs XLM-RoBERTa to encode target sub-words, optimizing with mean squared error and selecting the best checkpoint over ten epochs. Binary labels for Sub-task 1 are derived using a threshold of $\delta = 2$. The best-performing system, using Support Vector Machines (SVMs), achieved an $F_1$-macro score of 85.00.

\paragraph{Dataset Processing}
For this study, we use only the monolingual dataset, excluding the cross-lingual test set. The data is provided in JSONL format, so no transformation of its structure was necessary for this adaptation.

\paragraph{Adapted Structure for LLMs}
Each instance in the dataset is represented in JSONL format, with attributes including an identifier, the lemma of the target word, two sentences containing the word, the positions of the word in both sentences, and a label indicating whether the word has the same meaning in both sentences (1 for same meaning, 0 for different meaning). Below is an example instance:

\begin{lstlisting}
{
  "id": "delicatezza.noun.14",
  "lemma": "delicatezza",
  "sentence1": "Il presidente AZZOLLINI , stante la delicatezza dei temi oggetto del provvedimento , ritiene opportuno richiedere la predisposizione di una relazione tecnica .",
  "sentence2": "Montare la panna senza zucchero in modo che diventi molto solida , quindi unirla al composto avendo cura di mescolarla con molta delicatezza .",
  "start1": 36,
  "end1": 47,
  "start2": 129,
  "end2": 140,
  "label": 0
}
\end{lstlisting}

\paragraph{Dataset Splits}  
The adapted dataset is distributed as follows:
\begin{itemize}
    \item Train: \texttt{train.jsonl}, containing 2,805 instances.  
    \item Dev: \texttt{dev.jsonl}, containing 500 instances.  
    \item Test: \texttt{test.jsonl}, containing 500 instances.  
\end{itemize}  

\paragraph{License and Distribution}  
The dataset is publicly available and can be accessed on the Hugging Face platform.\footnote{\url{https://huggingface.co/datasets/evalitahf/word_in_context}}

\subsection{Textual Entailment (TE)}
The Recognizing Textual Entailment task (RTE) was proposed at Evalita 2009\footnote{\url{https://www.evalita.it/campaigns/evalita-2009/tasks/textual-entailment}} as the Italian version of the RTE series of task for English~\cite{Bos2009}. In this task, the input consists of two sentences: the \textit{text} (T) and the \textit{hypothesis} (H). The model's objective is to determine whether the meaning of the hypothesis is logically entailed by the text.

\paragraph{Examples}
Some examples for the Textual Entailment are reported in Table \ref{tab:te_examples}.



\setlength\tabcolsep{5pt}
\begin{table*}[h!]
\centering
    \begin{tabularx}{\textwidth}{XX}
    \toprule
    \textbf{Italian} & \textbf{Translated in English} \\
    \midrule
    \multicolumn{2}{>{\hsize=\dimexpr2\hsize+2\tabcolsep+\arrayrulewidth\relax}c}{\textbf{Example 1 (Entailed)}}\\
    \\
    \textbf{T:} Parla di attività nei panni di direttore commerciale e, dopo sei mesi, di direttore generale. & \textbf{T:} He talks about his activities as a sales manager and, after six months, as a general manager. \\
    \textbf{H:} Parla di attività di direttore commerciale e, dopo sei mesi, di direttore generale. & \textbf{H:} He talks about his activities as a sales manager and, after six months, as a general manager. \\
    \midrule
    \multicolumn{2}{>{\hsize=\dimexpr2\hsize+2\tabcolsep+\arrayrulewidth\relax}c}{\textbf{Example 2 (Not Entailed)}}\\
    \\
    \textbf{T:} Il primo acquisto immobiliare fu un terreno in via Alciati a Milano, per 190 milioni di lire. & \textbf{T:} The first real estate purchase was a plot of land in via Alciati in Milan, for 190 million lire. \\
    \textbf{H:} Il primo acquisto è un terreno in via Alciati a Milano. & \textbf{H:} The first purchase was a plot of land in via Alciati in Milan. \\
    \bottomrule
    \end{tabularx}
\caption{Examples from the Textual Entailment dataset.}
\label{tab:te_examples}
\end{table*}

\paragraph{Dataset} The dataset\footnote{\url{https://live.european-languagegrid.eu/catalogue/corpus/8121}} consists of pairs of texts taken from the revision histories of Italian Wikipedia articles. Each pair is labeled as either entailed or not entailed. The training set contains 400 pairs, equally divided between positive and negative examples, while the test set consists of 400 pairs, with 220 positive examples and 180 negative examples.\\

\paragraph{Evaluation Measure} The evaluation metric for this task is accuracy, defined as the ratio of correctly recognized pairs to the total number of pairs.\\

\paragraph{Results} A baseline system that predicts entailment for every example, regardless of the actual relationship between the pairs, yields an accuracy of 55\%. The top-performing system at Evalita 2009 exploited EDITS (Edit Distance Textual Entailment Suite), an open-source tool for Recognizing Textual Entailment (RTE), and reached an accuracy of 71\%.

\paragraph{Dataset Processing}
For this study, we use the original data provided in XML format, which has been transformed into JSONL format for easier accessibility and model processing. This transformation maintains the original structure of the dataset.

\paragraph{Dataset Structure}  
Each instance in the adapted dataset is represented in JSONL format with the following attributes:

\texttt{id}: Unique identifier for each instance.
\texttt{entailment}: Label showing whether the hypothesis is entailed by the text (e.g., ‘‘SI’’ for entailed, ‘‘NO’’ for not entailed).
\texttt{text1}: text.
\texttt{text2}: hypothesis.
An example instance is shown below:

\begin{lstlisting}
{
  "id": "1001",
  "entailment": "SI",
  "text1": "Pieralfonso Fratta Pasini e' un imprenditore e un politico italiano.",
  "text2": "Pieralfonso Fratta Pasini e' un imprenditore e politico italiano."
}
\end{lstlisting}

\paragraph{Dataset Splits}  
The adapted dataset is distributed as follows:
\begin{itemize}
    \item Dev: \texttt{valid.jsonl}, containing 400 instances.  
    \item Test: \texttt{test.jsonl}, containing 400 instances.  
\end{itemize}  

\paragraph{License and Distribution}  
The dataset is publicly available and can be accessed on the Hugging Face platform.\footnote{\url{https://huggingface.co/datasets/evalitahf/textual_entailment}}

\subsection{Sentiment Analysis (SA)}
The SENTIment POLarity Classification task (SENTIPOLC) was introduced as part of the Evalita 2016 campaign\footnote{\url{https://www.evalita.it/campaigns/evalita-2016/tasks-challenge/sentipolc}} and is described in detail in~\cite{Barbieri2016}. This task focuses on sentiment analysis of Italian tweets and includes three subtasks: polarity classification, subjectivity classification and irony detection. Among these, this study specifically focuses on polarity classification, where the objective is to categorize the sentiment of a tweet into one of four classes: positive, negative, neutral or mixed.

\paragraph{Examples}
Some examples for the Sentiment Analysis task are reported in Table \ref{tab:sa_examples}.


\setlength\tabcolsep{5pt}
\begin{table*}[h!]
\centering
    \begin{tabularx}{\textwidth}{XX}
    \toprule
    \textbf{Italian} & \textbf{Translated in English} \\
    \midrule
    \multicolumn{2}{>{\hsize=\dimexpr2\hsize+2\tabcolsep+\arrayrulewidth\relax}c}{\textbf{Positive}}\\
    \\
    Splendida foto di Fabrizio, pluri cliccata nei siti internazionali di Photo Natura & Fabrizio's splendid photo, clicked many times on Photo Natura's international sites \\
    \midrule
    \multicolumn{2}{>{\hsize=\dimexpr2\hsize+2\tabcolsep+\arrayrulewidth\relax}c}{\textbf{Negative}}\\
    \\
    Monti, ripensaci: l’inutile Torino-Lione inguaia l’Italia & Monti, think again: the useless Turin-Lyon is getting Italy into trouble \\
    \midrule
    \multicolumn{2}{>{\hsize=\dimexpr2\hsize+2\tabcolsep+\arrayrulewidth\relax}c}{\textbf{Neutral}}\\
    \\
    Primo passaggio alla strabrollo ma secondo me non era un iscritto & First pass at Strabrollo but in my opinion he wasn't a member \\
    \midrule
    \multicolumn{2}{>{\hsize=\dimexpr2\hsize+2\tabcolsep+\arrayrulewidth\relax}c}{\textbf{Mixed}}\\
    \\
    Dati negativi da Confindustria che spera nel nuovo governo Monti &   Negative data from Confindustria which hopes for the new Monti government \\
    \bottomrule
    \end{tabularx}
\caption{Examples from the Sentiment Analysis dataset.}
\label{tab:sa_examples}
\end{table*}

\paragraph{Dataset} The dataset for SENTIPOLC\footnote{\url{https://live.european-language-grid.eu/catalogue/corpus/7479}} comprises Italian tweets from various domains, including political and generic topics. The training dataset contains 7,410 examples, while the test dataset includes 2,000 examples.\\

\paragraph{Evaluation Measure} The evaluation of the sentiment classification task is done independently for positive and negative polarity. The overall $F_1$-macro score for the task is obtained by averaging the $F_{1}$ scores for both the positive and negative polarities. 

\paragraph{Results} A majority class baseline is used to establish a lower bound for performance, predicting the most frequent sentiment category for all tweets, resulting in an $F_1$-macro of 41.63. The best-performing model achieved an $F_1$-macro of 66.38, using deep learning techniques that represent individual tweets through geometric representations, such as vectors. The words in each tweet are represented as Word Embeddings, primarily generated using the Word2Vec tool.

\paragraph{Dataset Processing} 
For this study, we use the original dataset, which was provided in CSV format. It has been transformed into JSONL format for easier accessibility and processing. This transformation maintains the original structure of the dataset.

\paragraph{Dataset Structure}  
Each instance in the adapted dataset is represented in JSONL format with the following attributes:

\texttt{idtwitter}: Unique identifier for the tweet.
\texttt{subj}: Subjectivity label (1 for subjective, 0 for objective).
\texttt{opos}: Binary label for positive sentiment (1 for positive, 0 for non-positive).
\texttt{oneg}: Binary label for negative sentiment (1 for negative, 0 for non-negative).
\texttt{iro}: Binary label for irony (1 for ironic, 0 for non-ironic).
\texttt{lpos}: Label for low-level positive sentiment.
\texttt{lneg}: Label for low-level negative sentiment.
\texttt{top}: Text of the tweet.

An example instance is shown below:

\begin{lstlisting}
{"idtwitter": "133535736656953344", "subj": "1", "opos": "1", "oneg": "0", "iro": "0", "lpos": "1", "lneg": "0", "top": "1", "text": "Mario Monti sarebbe quello in grado di unire i due schieramenti nel dopo #Berlusconi. Mari e Monti insomma."}
\end{lstlisting}

\paragraph{Dataset Splits}  
The adapted dataset is distributed as follows: 
\begin{itemize}
    \item Train: \texttt{train.jsonl}, containing 7,410 instances.  
    \item Test: \texttt{test.jsonl}, containing 2,000 instances.  
\end{itemize}  

\paragraph{License and Distribution}  
The dataset is publicly available and can be accessed on the Hugging Face platform.\footnote{\url{https://huggingface.co/datasets/evalitahf/sentiment_analysis}}

\subsection{Hate Speech (HS)}
HaSpeeDe 2 at Evalita 2020\footnote{\url{http://www.di.unito.it/~tutreeb/haspeede-evalita20/index.html}} includes a Main Task and two Pilot Tasks, using datasets derived from Twitter messages and news headlines. The Main Task, which is the focus of this paper, consists in determining whether a given text contains hateful content directed at specific target groups, namely immigrants, Muslims, and Roma.

\paragraph{Examples}
Some examples for the Hate Speech task are reported in Table \ref{tab:hs_examples}.


\setlength\tabcolsep{5pt}
\begin{table*}[h!]
\centering
    \begin{tabularx}{\textwidth}{XX}
    \toprule
    \textbf{Italian} & \textbf{Translated in English} \\
    \midrule
    \multicolumn{2}{>{\hsize=\dimexpr2\hsize+2\tabcolsep+\arrayrulewidth\relax}c}{\textbf{Hate Speech}}\\
    \\
    Via tutti i campi Rom e disinfettare per bene il lerciume che si lasciano dietro. & Remove all the Roma camps and thoroughly disinfect the filth they leave behind. \\
    \midrule
    \multicolumn{2}{>{\hsize=\dimexpr2\hsize+2\tabcolsep+\arrayrulewidth\relax}c}{\textbf{Not Hate Speech}}\\
    \\
    Sea Watch, Finanza sequestra la nave: sbarcano i migranti. & Sea Watch, Finance seizes the ship: migrants disembark. \\
    \bottomrule
    \end{tabularx}
\caption{Example from the Hate Speech dataset.}
\label{tab:hs_examples}
\end{table*}

\paragraph{Dataset} The HaSpeeDe2 dataset\footnote{\url{https://live.european-language-grid.eu/catalogue/corpus/7498/overview}} collects 6,839 tweets for the training dataset, 1,263 test tweets, and 500 test news headlines, all annotated for hate speech detection.

\paragraph{Evaluation Measure} $F_1$-macro is used to evaluate systems, which balances precision and recall across classes.\\

\paragraph{Results} The baseline models consisted of an SVM-based approach and a Most Frequent Class (MFC) model. For Twitter data, the SVM baseline scored 72.12, while the MFC model performed much worse with a score of 33.66. On News data, the SVM baseline achieved a score of 62.10, and the MFC model scored 38.94. The top-performing systems were based on BERT and its derivative architectures. These systems achieved a $F_1$-macro score of 80.88 on Twitter data, while the highest score on News data was 77.44. 

\paragraph{Dataset Processing} 
The dataset is originally provided in TSV format. For ease of use, it has been converted to JSONL format, with the original structure of the data preserved.

\paragraph{Dataset Structure}
An example instance from the dataset in JSONL format is shown below:  

\begin{lstlisting}
{"id": "4968", "full_text": "Ferocissimo Feltri: 'Gli islamici ci odiano! Tutti fuori dai coglioni, rimandiamoli tra i cammelli' URL via @user", "hs": 1, "stereotype": 1}
\end{lstlisting}

\paragraph{Dataset Splits}
The dataset is divided as follows:  
\begin{itemize}
    \item Train: \texttt{hs\_dev.jsonl}, containing 6,839 instances.  
    \item Test: \texttt{hs\_test\_all.jsonl}, containing a total of 1,763 instances, combining both 1,263 tweets and 500 news headlines.  
\end{itemize}  

\paragraph{License and Distribution}
The dataset is publicly available and can be accessed on the Hugging Face platform.\footnote{\url{https://huggingface.co/datasets/evalitahf/hatespeech_detection}}  

\subsection{Frequently Asked Questions \& Question answering (FAQ)}
The QA4FAQ task was part of the Evalita 2016 campaign\footnote{\url{https://www.evalita.it/campaigns/evalita-2016/tasks-challenge/qa4faq}}. It focuses on retrieving the most relevant Frequently Asked Questions (FAQs) and their corresponding answers based on a user query. Given a user query, the participating systems must search the FAQ database to find the most similar question and return the answer to that question. Detailed information about the task can be found in~\cite{Caputo2016}. This task evaluates the performance of Question Answering (QA) and Information Retrieval (IR) systems in scenarios where the document collection consists entirely of FAQs.

\paragraph{Examples}
Some examples for the FAQ Task are reported in Table \ref{tab:faq_examples}.


\setlength\tabcolsep{5pt}
\begin{table*}[h!]
\centering
    \begin{tabularx}{\textwidth}{XX}
    \toprule
    \textbf{Italian} & \textbf{Translated in English} \\
    \midrule
    \multicolumn{2}{>{\hsize=\dimexpr2\hsize+2\tabcolsep+\arrayrulewidth\relax}c}{\textbf{Example 1 (Entailed)}}\\
    \\
    \textbf{Q:} Come posso telefonare al numero verde da un cellulare? & \textbf{Q:} How can I call the toll-free number from a mobile phone? \\
    \textbf{A:} È possibile chiamare il Contact Center AQP per segnalare un guasto o per un pronto intervento telefonando gratuitamente anche da cellulare al numero verde 800.735.735. & \textbf{A:} You can call the AQP Contact Center to report a fault or for emergency assistance by calling the toll-free number 800.735.735 free of charge even from a mobile phone.\\
    \bottomrule
    \end{tabularx}
\caption{Example from the FAQ dataset. In this example, the FAQ is relevant to the user query: \textit{Si può telefonare da cellulare al numero verde?} The system retrieves the most similar FAQ question to the user query and provides the corresponding answer.}
\label{tab:faq_examples}
\end{table*}

\paragraph{Dataset} The dataset consists of a knowledge base with 406 FAQs, each containing a question, an answer, and associated tags. It also includes 1,132 user queries extracted from the AQP Risponde system logs. A set of 1,406 query-FAQ relevance pairs, derived from user feedback, was manually checked for quality by AQP customer support experts.

\paragraph{Evaluation Measure} Participants' systems are ranked according to accuracy@1, which measures precision based on the first answer provided.

\paragraph{Results} A baseline system using Apache Lucene was provided. It indexed FAQ fields (question, answer, and tags) and used the ItalianAnalyzer for retrieval, with boosted relevance for each field. The baseline achieved a c@1 score of 40.76. The best-performing system used a cognitive QA model with field-specific similarity measures and query expansion via Wiktionary, achieving a c@1 score of 44.39.

\paragraph{Dataset Processing} 
To create a new multiple-choice task derived from QA4FAQ, the dataset was transformed. Each question in the original dataset is now paired with four possible answers, one of which is correct.

\paragraph{Dataset Structure}  
An example instance from the dataset in JSONL format is shown below:  

\begin{lstlisting}
{
  "id": "258", 
  "question": "Quale uso e' previsto per un locale commerciale temporaneamente vuoto?", 
  "correct_answer": "D", 
  "A": "Dipende principalmente dalle quote altimetriche: le zone alte degli abitati sono spesso piu' critiche di altre perche' a parita' di pressione in rete occorre superare un maggiore 'dislivello' per servire le abitazioni. In pratica, nelle zone alte degli abitati, in caso di riduzione di pressione, la rete fatica a riempirsi ed a garantire un livello ottimale di servizio.", 
  "B": "E' possibile effettuare il subentro e/o la modifica contrattuale presentandosi personalmente presso gli uffici commerciali provvisti di dati e documentazione richiesti e matricola del contatore, lettura del contatore aggiornata. In assenza di morosita', il contratto sara' attivato in tempo reale.", 
  "C": "Si', la richiesta deve essere formulata in modo tale da prevedere anche l'allacciamento relativo alla fogna e depurazione.", 
  "D": "L'utenza sara' ad uso domestico e il contratto intestato al proprietario"
}
\end{lstlisting}

\paragraph{Dataset Splits}  
The dataset is split as follows:  
\begin{itemize}
    \item Trial: \texttt{multichoice\_v1\_test.jsonl}, containing 5 instances (for 5-shot learning tasks). 
    \item Test: \texttt{multichoice\_v1\_dev.jsonl} containing 401 instances 
\end{itemize}  

\paragraph{License and Distribution}  
The dataset is publicly available and can be accessed on the Hugging Face platform.\footnote{\url{https://huggingface.co/datasets/evalitahf/faq}}

\subsection{Admission Tests (AT)}
This task is detailed in the study~\cite{Casola2023} and is not part of the Evalita evaluation campaign. It involves answering multiple-choice questions from the Italian medical specialty exams, where each question has five answer options, with only one being correct. The questions assess knowledge across various medical domains, including clinical reasoning and diagnosis, and some questions require deeper analysis beyond basic factual knowledge.\\

\paragraph{Example:}
Some examples for the Admission Test task are reported in Table \ref{tab:at_examples}.

\setlength\tabcolsep{5pt}
\begin{table*}[h!]
\centering
    \begin{tabularx}{\textwidth}{XX}
    
    \toprule
    \textbf{Italian} & \textbf{Translated in English} \\
    \midrule
    \textbf{Domanda:} Un paziente di 49 anni in corso di induzione di anestesia generale per colecistectomia presenta bradicardia grave.  Che cosa è indicato somministrare? & \textbf{Question:} A 49-year-old patient undergoing induction of general anesthesia for cholecystectomy presents with severe bradycardia. What is indicated to administer?\\
   
    \parbox{.45\textwidth}{
        \begin{list}{}{\leftskip=2em} 
              \item A. Betametasone
              \item B. Atenololo
              \item C. Propofol
              \item D. Fentanil
              \item E. Atropina
        \end{list}
    } & \parbox{.45\textwidth}{
        \begin{list}{}{\leftskip=2em} 
            \item A. Betamethasone
            \item B. Atenolol
            \item C. Propofol
            \item D. Fentanyl
            \item E. Atropine
        \end{list}
    }\\ [1cm]
    La risposta corretta è E, Atropina &
    The correct answer is E, Atropine \\
    \bottomrule
    \end{tabularx}
\caption{Example from the AT dataset.}
\label{tab:at_examples}
\end{table*}

\paragraph{Original Dataset} The dataset is originally provided in Excel format and includes 136 multiple-choice questions (after excluding 4 that required images, such as ECGs or medical images). These questions were extracted from the 2022 entrance exams for the Italian Medical Specialty Schools (SSM). Each question has five possible answers, with one correct answer. The questions span a wide range of medical topics, including diagnosis, clinical treatment protocols, and reasoning. Since the task is based on the complete set of 136 questions, there is no formal training phase for the model; instead, evaluation is conducted by directly presenting the test questions to the model.

\paragraph{Evaluation Measure} Model performance was evaluated based on accuracy.

\paragraph{Results} The GPT-3.5-turbo model achieved an average accuracy of 77.14\% on this dataset.


\paragraph{Dataset Processing} 
To ensure a fair evaluation, we expanded the original dataset, which contained only 136 examples, to a more comprehensive version with 521 instances.
The additional data was collected by selecting ATs from different years.
As a result, the dataset\footnote{\url{https://huggingface.co/datasets/evalitahf/admission_test/tree/main}} now includes SSM entrance exams for the years 2017, 2018, 2019, in addition to the 2022 exams from the original dataset.
Following the approach of the original dataset, we excluded all questions that required image interpretation.

All data were converted to the JSONL format. During this transformation, the answer choices were shuffled to prevent the correct answer from consistently appearing in the first position, as the original dataset always listed it as option A.
Each instance in the transformed dataset consists of a medical question along with its associated multiple-choice answers.

\paragraph{Adapted Structure for LLMs}  
An example instance from the dataset in JSONL format is shown below:\\

\begin{lstlisting}
{
  "Id": 1,
  "Question": "Un uomo di 23 anni, dopo una cena a base di crostacei, presenta pomfi localizzati su tutto il corpo e prurito diffuso. Riferisce inoltre comparsa di dispnea e una sensazione di nodo alla gola. Tutti i farmaci riportati potrebbero avere un ruolo nella gestione di questa reazione, TRANNE uno: quale?",
  "A": "Salbutamolo",
  "B": "Adrenalina",
  "C": "Corticosteroidi sistemici",
  "D": "Anti-istaminici H1 ",
  "E": "Beta-bloccante ",
  "Correct": "E"
}
\end{lstlisting}

\paragraph{Adapted Dataset Splits}  
The dataset is divided as follows:  
\begin{itemize}
    \item Trial: \texttt{trial.jsonl}, containing 21 instances (for 5-shot learning tasks).  
    \item Test: \texttt{test.jsonl}, containing 500 instances.  
\end{itemize}  

\paragraph{License and Distribution}  
The dataset is publicly available and can be accessed on the Hugging Face platform.\footnote{\url{https://huggingface.co/datasets/evalitahf/admission_test}}

\subsection{Lexical Substitution (LS)}
Task A of the Lexical Substitution task at Evalita 2009\footnote{\url{https://www.evalita.it/2009/tasks/lexical}} focuses on replacing a target word in a specific context with its most suitable synonyms in Italian, as described in~\cite{Toral2009}. In particular, the task requires systems to provide the lemma of the synonyms that are contextually relevant, without relying on a predefined sense inventory. This allows both supervised and unsupervised approaches to be applied.

\paragraph{Example}
Some examples of the Lexical Substitution task are reported in Table \ref{tab:ls_examples}.
\setlength\tabcolsep{5pt}
\begin{table*}[h!]
\centering
    \begin{tabularx}{\textwidth}{XX}
    
    \toprule
    \textbf{Italian} & \textbf{Translated in English} \\
    \midrule
    \textbf{Context 1:} Ma 8 milioni nel 1939 non sono una cifra \textit{modesta}. &
    \textbf{Context 1:} But 8 million in 1939 is not a \textit{modest} figure. \\ 
    \textbf{Substitutions:} esiguo, scarso, piccolo, modico & 
    \textbf{Substitutions:} meager, scarce, small, moderate \\ 
    \\
    \hline 
    \textbf{Context 2:} Ma Nanni non ha mai avuto grandi capacità formali alla Lang o alla Kubrick (le cui opere sono disperatamente segnate dai loro autori), i suoi film sono fatti di immagini programmaticamente \textit{modeste}.  &
    \textbf{Context 2:} But Nanni has never had great formal skills a la Lang or Kubrick (whose works are desperately marked by their authors), his films are made of programmatically \textit{modest} images. \\ 
    \textbf{Substitutions:} semplice & \textbf{Substitutions:} simple \\
    \bottomrule
    \end{tabularx}
\caption{Examples from the LS dataset.}
\label{tab:ls_examples}
\end{table*}

\paragraph{Original Dataset} The original dataset for this task includes 2,010 annotated contexts, containing 231 target words (75 nouns, 58 adjectives, 63 verbs, and 36 adverbs) derived from language resources such as ItalWordNet. The data is split into 300 contexts for a trial set and 1,710 for a test set.

\paragraph{Evaluation Measure} Systems are evaluated using two scoring types: \textit{Best}, which measures precision, recall, and $F_1$ measure of the top synonym proposed, and \textit{Out-of-Ten (oot)}, which considers the top 10 guesses.

\paragraph{Results} The baseline system uses semantic relations from ItalWordNet (IWN) and PAROLE-SIMPLE-CLIPS (PSC). The baseline based on PSC outperformed all participant systems for the \textit{Best} score type, achieving an $F_1$ measure of 9.88. The baseline that combined IWN and PSC reached an F-measure of 25.19 for the \textit{oot} score. The top-performing system, on the other hand, constructs n-grams from the contexts where the target word is replaced by synonyms from a lexicon. These contexts are then searched in a corpus, and the substitutes are weighted based on various factors, including the number of documents retrieved. This system achieved an F-measure of 7.64 for the \textit{Best} score and 38.82 for the \textit{oot} score. 



\paragraph{Dataset Processing} 
Despite \cite{Toral2009} stating that the trial set was composed of 300 examples, we were only able to obtain a version of 272 instances.
The dataset is already in JSONL format, and no transformations were needed. 

\paragraph{Adapted Structure for LLMs}  
An example instance from the dataset in JSONL format is shown below:  

\begin{lstlisting}
{"id": "301", 
"context": "Ma in giro per la provincia di Vicenza esistono industrie che lavorano con <head>sostanze</head> ben piu' pericolose : che conseguenze avrebbe un attentato contro queste aziende ?", 
"head": "sostanze", 
"lexelt": "sostanza.n", 
"lexelt_lemma": "sostanza", 
"answers": [{"word": "materiale", "count": 2}, {"word": "materia", "count": 2}, {"word": "composto", "count": 1}]}
\end{lstlisting}

\paragraph{Adapted Dataset Splits}  
The dataset is split as follows:  
\begin{itemize}
    \item Trial:\texttt{trial.jsonl}, containing 272 instances. 
    \item Test: \texttt{test\_v2.jsonl}, containing 1,710 instances. 
\end{itemize}  

\paragraph{Stop Tokens for Generation}
The characters used as stop tokens for the \textit{generate\_until} method are: $<\!/s\!>$.

\paragraph{License and Distribution}  
The dataset is publicly available and can be accessed on the Hugging Face platform.\footnote{\url{https://huggingface.co/datasets/evalitahf/lexical_substitution}}

\subsection{Named Entity Recognition (NER)}
Named Entities Recognition on Multi-Domain Documents task at Evalita 2023\footnote{\url{https://nermud.fbk.eu}} focuses on identifying and classifying Named Entities (NEs) in diverse Italian text domains \cite{Aprosio2023}. The shared task is divided into two sub-tasks: domain-agnostic classification, using a single model across domains, and domain-specific classification, leveraging separate models tailored to specific text types such as news, literature, and political speeches.

\paragraph{Example}
Some examples for the Named Entities Recognition task are reported in Table \ref{tab:ner_examples}.

\setlength\tabcolsep{5pt}
\begin{table*}[h]
\centering
    \begin{tabularx}{\textwidth}{XX}
     \textbf{Italian} & \textbf{Translated in English} \\
    \toprule
   
    L'astronauta [B-PER Umberto] [I-PER Guidoni], dell' [B-ORG Agenzia] [I-ORG Spaziale] [I-ORG Europea], svela ai bambini i segreti della [B-LOC Stazione] [I-LOC Spaziale] [I-LOC Internazionale]. & 
    The astronaut [B-PER Umberto] [I-PER Guidoni], from the [B-ORG European] [I-ORG Space] [I-ORG Agency], reveals the secrets of the [B-LOC International] [I-LOC Space] [I-LOC Station]. \\
    \midrule
    
    \end{tabularx}
\caption{Example from the NER dataset.}
\label{tab:ner_examples}
\end{table*}

\paragraph{Original Dataset}
The dataset comprises part of the Kessler Italian Named-entities Dataset (KIND) amounting to over 700,000  tokens. It features annotations for person, organization, and location entities from three domains:
\begin{itemize}
\item Wikinews (WN): 1,198 articles (364,816 tokens) from the last 20 years.
\item Literature (FIC): 86 chapters from public domain books (219,638 tokens).
\item Political Writings (ADG): 173 documents (164,537 tokens) by Alcide De Gasperi.
\end{itemize}

\paragraph{Evaluation Measure}
$F_1$-macro scores evaluate system performance on person, location, and organization entities. Results are also analyzed per domain to assess system robustness across text types.

\paragraph{Results}
Two baselines were provided:
\begin{itemize}
\item CRF with gazetteers: Achieved an overall $F_1$ of 0.83.
\item BERT implementation: Achieved an overall $F_1$ of 0.89.
\end{itemize}

The best performing system used instruction-tuned models based on IT5 and LLaMA, achieving an $F_1$-macro of 0.88 with their best configuration, which matched the baselines in most cases.

\paragraph{Dataset Processing} 
The original dataset was sentence-split and transformed into JSON format, where each instance corresponds to a sentence and includes the annotated entities occurring in that sentence.

\paragraph{Adapted Structure for LLMs}  
An example instance from the dataset in JSON format is shown below:  

\begin{lstlisting}
{
  "text": "Altra difficolta' consisteva, riguardo alle aspirazioni nazionali su Roma, nell' attitudine di Napoleone III, imperatore dei Francesi, il quale, ora che Cavour era morto, pur riconoscendo il Regno d' Italia, aveva dichiarato che dava valore alle proteste del Papa contro l' occupazione da noi fatta di parecchie delle province pontificie e che percio' continuava a occupare Roma per difenderla da possibili invasioni.",
  "entities": [
    {"entity_text": "Roma", "type": "LOC"},
    {"entity_text": "Napoleone III", "type": "PER"},
    {"entity_text": "Cavour", "type": "PER"},
    {"entity_text": "Regno d' Italia", "type": "ORG"},
    {"entity_text": "Roma", "type": "LOC"}
  ]
}
\end{lstlisting}

\paragraph{Adapted Dataset Splits}  
The dataset retains the original split across the three domains: Wikinews (WN), Literature (FIC), and Political Writings (ADG).
\begin{itemize}
    \item \textbf{ADG Domain:}  
        \begin{itemize}
            \item \texttt{adg\_train}: 5,147 instances  
            \item \texttt{adg\_dev}: 1,122 instances  
            \item \texttt{adg\_trial}: 5 instances (for 5-shot learning tasks)  
            \item \texttt{adg\_test}: 521 instances  
        \end{itemize}
    \item \textbf{FIC Domain:}  
        \begin{itemize}
            \item \texttt{fic\_train}: 11,423 instances  
            \item \texttt{fic\_dev}: 1,051 instances  
            \item \texttt{fic\_trial}: 5 instances (for 5-shot learning tasks)  
            \item \texttt{fic\_test}: 1,517 instances  
        \end{itemize}
    \item \textbf{WN Domain:}  
        \begin{itemize}
            \item \texttt{wn\_train}: 10,912 instances  
            \item \texttt{wn\_dev}: 2,594 instances  
            \item \texttt{wn\_trial}: 5 instances (for 5-shot learning tasks)  
            \item \texttt{wn\_test}: 2,088 instances  
        \end{itemize}
\end{itemize}

\paragraph{Stop Tokens for Generation}
The characters used as stop tokens for the \textit{generate\_until} method are: $<\!/s\!>$, /n.

\paragraph{License and Distribution}  
The dataset is publicly available and can be accessed on the Hugging Face platform.\footnote{\url{https://huggingface.co/datasets/evalitahf/entity_recognition}}

\subsection{Relation Extraction (REL)}
The CLinkaRT task at Evalita 2023\footnote{\url{https://e3c.fbk.eu/clinkart}} is a relation extraction task focused on the clinical domain, specifically addressing the association between laboratory results (RML) and their respective test EVENTS in Italian clinical narratives, as described in the CLinkaRT overview \cite{Altuna2023}. In this context, RML refers to the result or value of a laboratory test, such as a numerical value or range, while EVENT represents the corresponding laboratory test or measurement. Participants used clinical cases, each represented by both the document text and its annotated relations, to implement and evaluate systems for this task. Every annotated relation is presented on a separate line and is represented as an ordered pair of entity mentions (i.e., RML, event), with each entity mention defined by its start and end character offsets.

\paragraph{Example}
Some examples for the Relation Extraction task are reported in Table \ref{tab:rel_examples}.

\setlength\tabcolsep{5pt}
\begin{table*}[h!]
\centering
    \begin{tabularx}{\textwidth}{XX}
    
    \toprule
    \textbf{Italian} & \textbf{Translated in English} \\
    \midrule
     \textbf{Document:} 100854|t|Un maschio di 25 anni dopo una partita di calcetto si è presentato in PS per paralisi degli AAII. Gli ematochimici evidenziavano marcata ipokaliemia: $<$1.5 Mmol/L; dopo esecuzione di supplementazione di potassio il pz veniva dimesso. Pochi giorni più tardi il pz si ripresentava in PS per una recidiva e veniva ricoverato. Veniva eseguito TSH con riscontro di tireotossicosi senza sintomi di ipertiroidismo: TSH $<$0.005 mUI/L, FT4 44.9 ng/L. &
    \textbf{Document:} 100854|t| A 25-year-old male after a soccer game presented to the emergency department for AAII paralysis. Hematochemicals showed marked hypokalemia: $<$1.5 Mmol/L; after performing potassium supplementation the patient was discharged. A few days later the patient reappeared in the emergency department for a recurrence and was admitted. TSH was performed with thyrotoxicosis found without symptoms of hyperthyroidism: TSH $<$0.005 mUI/L, FT4 44.9 ng/L. \\ 
    \\
    \parbox{.45\textwidth}{
        \textbf{Relations:}
        \begin{itemize} 
            \item 100854     REL     150-161   137-148   $<$1.5 Mmol/L   ipokaliemia
            \item 100854     REL     411-423   407-410   $<$0.005 mUI/L  TSH
            \item 100854     REL     429-438   425-428   44.9 ng/L     FT4
        \end{itemize}
        
    } & \parbox{.45\textwidth}{
        \textbf{Relations:}
        \begin{itemize} 
            \item 100854     REL     150-161   137-148   $<$1.5 Mmol/L   ipokaliemia
            \item 100854     REL     411-423   407-410   $<$0.005 mUI/L  TSH
            \item 100854     REL     429-438   425-428   44.9 ng/L     FT4
        \end{itemize}
        
    } \\
    
    \bottomrule
    \end{tabularx}
\caption{Example from the REL dataset.}
\label{tab:rel_examples}
\end{table*}

\paragraph{Original Dataset} The dataset is derived from the E3C corpus \cite{Magnini2020} and is split into training, development, and test sets. It includes 83 clinical cases (training/development) and 80 cases (test), annotated with relations linking results (RML) to tests (EVENT).  

\paragraph{Evaluation Measure}  
Performance is assessed using Precision, Recall, and $F_1$ score.

\paragraph{Results} A fine-tuned mBERT model, serving as a baseline, achieved an $F_1$ of 62.83, whereas GPT-based few-shot learning, employed as an unsupervised baseline, performed notably worse with an $F_1$ of 36.79. The best-performing system achieved an $F_1$ of 62.99. Unlike traditional approaches that separately extract entities and relations, this method identifies EVENTS first, followed by the creation of relations. In their study, various BERT-based models were evaluated, including Italian BERT and MedBIT-R3-plus. The latter, specifically pre-trained for the medical domain, achieved the best performance following fine-tuning on an augmented dataset.

\paragraph{Dataset Processing} 
A single transformation was applied to this dataset: sentences were extracted from each document based on the provided sentence boundaries. This transformation ensures that related entity mentions always occur within the same sentence and never span across multiple sentences. Consequently, each entry in the transformed dataset corresponds to a sentence, representing an instance of the dataset, along with a list of related entity mention pairs within that sentence.

\paragraph{Adapted Structure for LLMs} An example instance from the dataset is shown below:\\\\

\begin{lstlisting} 
{ "text": "Gli esami colturali (germi comuni, BK) risultavano negativi", "relations": [ ["negativi", "esami"], ["negativi", "BK"], ["negativi", "germi"] ] } 
\end{lstlisting}

\paragraph{Adapted Dataset Splits}
The dataset is divided as follows:
\begin{itemize} 
\item Train: \texttt{train\_v2.jsonl}, containing 1,113 sentences.
\item Trial: \texttt{trial\_v2.jsonl}, containing 5 sentences (for 5-shot learning tasks).
\item Test: \texttt{test\_v2.jsonl}, containing 1,064 sentences.
\end{itemize}

\paragraph{Stop Tokens for Generation}
The characters used as stop tokens for the \textit{generate\_until} method are: $<\!/s\!>$.

\paragraph{License and Distribution} The dataset is publicly available and can be accessed on the Hugging Face platform.\footnote{\url{https://huggingface.co/datasets/evalitahf/relation_extraction}}

\subsection{Summarization (SUM)}

The task of text summarization for the Italian language \cite{Landro2022} is addressed by the Fanpage and IlPost datasets \cite{Landro2022}. These resources were created to train models capable of generating concise summaries of news articles and other text content in Italian. Notably, this effort is independent of the Evalita evaluation campaign and aims to advance the field of Italian text summarization.

\paragraph{Example}
Some examples for the Summarization task are reported in Table \ref{tab:sum_examples}.
\setlength\tabcolsep{5pt}
\begin{table*}[h!]
\centering
    \begin{tabularx}{\textwidth}{XX}
    
    \toprule
    \textbf{Italian} & \textbf{Translated in English} \\
    \midrule
     \textbf{Document:} La serie tv The Big Bang Theory finirà nel 2019. Warner Bros. Television e CBS, che producono e trasmettono la serie, hanno annunciato che la serie terminerà con la 12ª stagione, il cui primo episodio sarà trasmesso negli Stati Uniti il 24 settembre e il cui ultimo episodio andrà in onda l’anno prossimo. In totale, quando sarà finita, The Big Bang Theory sarà composta da 279 episodi e sarà la più lunga sitcom multi-camera (diversa quindi da quelle single-camera, come Scrubs o Modern Family) della storia della tv statunitense. Il primo episodio ... Sempre su Infinity si può vedere anche Young Sheldon, una serie prequel e spin-off che racconta la vita di Sheldon Cooper da giovane. & 
    \textbf{Document:} The TV series The Big Bang Theory will end in 2019. Warner Bros. Television and CBS, which produce and broadcast the series, have announced that the series will end with the 12th season, the first episode of which will air in the United States on Sept. 24 and the last episode of which will air next year. In total, when it is finished, The Big Bang Theory will consist of 279 episodes and will be the longest-running multi-camera sitcom (thus different from single-camera sitcoms such as Scrubs or Modern Family) in U.S. TV history. The first episode ... Also on Infinity is Young Sheldon, a prequel and spin-off series about Sheldon Cooper's life as a young man.\\
    \\
    \textbf{Summary:} The Big Bang Theory finirà nel 2019. Non è stata rinnovata e terminerà con la 12ª stagione, dopo 279 episodi. &
    \textbf{Summary:} The Big Bang Theory will end in 2019. It has not been renewed and will end with season 12, after 279 episodes. \\
    \bottomrule
    \end{tabularx}
\caption{Example from the Summarization dataset.}
\label{tab:sum_examples}
\end{table*}

\paragraph{Original Dataset}
The Fanpage dataset includes 84,308 Italian news articles spanning 9 categories. The articles are structured with titles, brief summaries, and detailed descriptions. The IlPost dataset is smaller with 44,025 news articles.

\paragraph{Evaluation Measure}
Performance is evaluated using ROUGE scores (ROUGE-1, ROUGE-2, ROUGE-L, ROUGE-LS), which measure the quality of generated summaries in terms of overlap with reference summaries.

\paragraph{Results} For the Fanpage dataset, the best-performing system was mBART, achieving the highest ROUGE scores across the board: ROUGE-1: 36.50 and ROUGE-2: 17.44. For the IlPost dataset, mBART again outperformed the other systems with the best ROUGE scores: ROUGE-1: 38.91 and ROUGE-2: 21.38.

\paragraph{Dataset Processing} 
For evaluation purposes, we used a random selection of 100 instances from the Fanpage dataset. The original data was provided in CSV format, which was then transformed into JSONL format for better accessibility, maintaining the original structure. 

\paragraph{Adapted Structure for LLMs}  
An example instance from the dataset in JSONL format is shown below:  

\begin{lstlisting}
{"source": "Cara Europa ti scrivo. Il Presidente del Consiglio Silvio Berlusconi ha inviato a Bruxelles la tanto attesa lettera. Attesa dalle istituzioni europee, dalla cancelliera tedesca Angela Merkel e dal premier francese Nicolas Sarkozy, ma soprattutto dagli italiani che aspettano di capire come si muovera' l'esecutivo per tappare le falle di una nave che imbarca acqua. ... L'ipotesi potrebbe essere quella di anticipare al 2012 (prima era stato stabilito il 2014) il percorso graduale dell'eta' pensionabile da 60 a 65 anni delle donne del settore privato.", 
"target": "Nella missiva inviata a Bruxelles, il Cavaliere avrebbe parlato dell'obiettivo del pareggio di bilancio nel 2013, del decreto sviluppo e della riforma delle pensioni. "}
\end{lstlisting}

\paragraph{Adapted Dataset Splits}
The dataset is divided as follows:
\begin{itemize} 
\item Train: \texttt{val.jsonl}, containing 8,436 instances (10\% of the Fanpage dataset).
\item Test: \texttt{test\_100.jsonl}, containing 100 instances.
\end{itemize}

\paragraph{Stop Tokens for Generation}
The characters used as stop tokens for the \textit{generate\_until} method are: $<\!/s\!>$.

\paragraph{License and Distribution}
The Fanpage dataset is publicly available and can be accessed on the Hugging Face platform.\footnote{\url{https://huggingface.co/datasets/evalitahf/summarization-fp}}

\section{Prompting}\label{sec:prompting}
The definition of the prompt to be used by LLMs for a task is crucial and sensitive \cite{liu2021pretrainpromptpredictsystematic} \cite{qin-eisner-2021-learning} \cite{sane2024notsosimplewaybeat}, as LLMs can be highly responsive to different prompt styles \cite{mizrahi2024stateartmultipromptllm} \cite{polo2024efficientmultipromptevaluationllms}. We mainly target instruct LLMs, which have been instructed to solve a number of conversational tasks, including several question answering situations. As a consequence, we expect that most LLMs are familiar with the tasks of the Evalita-LLM benchmark.

\subsection{General Guidelines for Prompting}
As the Evalita datasets do not come with associate prompts, we adopt the following general guidelines for prompt definition:

\begin{itemize}
    \item \textit{Language}. The whole prompt has to be in Italian, including labels that the model has to predict. This is because Evalita-LLM is intended as a benchmark for Italian, and using English prompts might introduce noise, particularly for generative tasks (see, for instance, \cite{marchisio-etal-2024-understanding}).
    \item \textit{Role of the model}. We avoid pre-pending instructions about the role of the model, like “you are an assistant…”). As different LLMs might be instructed to converse with specific prompts, we decided not to use them. 
    \item \textit{Minimality}. Minimally verbose prompts are preferred in order to reduce potential biases for a model. We use a simple and as short as possible wording.
    \item \textit{Input text}. After some tests, we decided to specify in the prompt the type of input text required by a  task. For instance, we ask the model to perform a certain task on a tweet, news, sentence, words, etc.
\end{itemize}

In addition to general guidelines for prompting, we adopt a compositional approach, identifying four possible components for our prompts: (i) A core question or statement mentioning the input for the task and what is expected. This element is mandatory in all prompts. (ii) A short description of the task. This element is optional and, if present, appears at the beginning of the prompt. (iii) For the multiple-choice setting, the possible options for the answers. This element is optional, and, if present, appears at the end of the prompt. (iv) For the generative setting, instructions about the output format. This element is optional, and, if present, appears at the end of the prompt.

The four elements composing a prompt are then combined to obtain the actual prompts for a task. After a number of tests, we have identified six prompts templates for multiple-choice tasks, and four prompts templates for generative tasks. The difference is mainly due to the high computation effort of the generative tasks.

\subsection{Prompts for Multiple-choice Tasks}

 As for multiple-choice tasks, we use the following six prompt templates, which are adapted to each task. 

\begin{itemize}
    \item \textit{Prompt 1: Question.} This prompt is formulated as a base question that the model must answer. The formulation should follow the general guidelines for prompt definition. As an example (see Table  \ref{tab:sentiment_prompts}), for a Sentiment Analysis task, the base question is \textit{Qual è il sentiment espresso nel seguente tweet '\{\{text\}\}'?}, where the variable '\{\{text\}\}' will be substituted by the tweets in the test data.
    \item \textit{Prompt 2: Task description + question.} In this prompt a short task description is pre-pended to the base question. As an example (see Table  \ref{tab:sentiment_prompts}), for a Sentiment Analysis task, the task description \textit{Devi svolgere un compito di analisi del sentiment.} is added before the base question reported for Prompt 1.
    \item \textit{Prompt 3: Question + answer.} In this prompt, the possible answers are appended to the base question. As an example (see Table  \ref{tab:sentiment_prompts}), for a Sentiment Analysis task, the answers \textit{A: Positivo \textbackslash n B: Negativo \textbackslash n C: Neutro \textbackslash n D: Misto} are appended after the base question reported in Prompt 1.
    \item \textit{Prompt 4: Task description + question + answer.} In this prompt we merge the previous prompts, pre-pending the task description and appending the answers to the base question.
    \item \textit{Prompt 5: Affermative.} This prompt is formulated as a simple affermative statement, without mentioning the possible answers. For instance, the following is the affermative statement used for Sentiment Analysis: \textit{Il seguente tweet: '\{\{text\}\}' esprime un sentiment}. 
    \item \textit{Prompt 6: Task description + affermative.} In this prompt the task description, the same used in Prompt 2, is pre-pended to the base affermative statement.
\end{itemize}

Table \ref{tab:sentiment_prompts} reports the six prompts used for the Sentiment Analysis task. Notice that, for multiple-choice prompts, answers can be either part of the prompt, and then just mentioned in the list of the options (as in Prompts 3 and 4 in Table \ref{tab:sentiment_prompts}), or they can be mentioned only in the options, as in Prompts 1 and 2 in Table \ref{tab:sentiment_prompts}.

\begin{table}[ht]
\centering
\renewcommand{\arraystretch}{1.5} 
\begin{tabular}{c|p{2cm}|p{6cm}|p{2cm}}
\toprule
\textbf{} & \textbf{Pattern} & \textbf{Prompt} & \textbf{Options} \\ \midrule
p1 & question & Qual è il sentiment espresso nel seguente tweet: '\{\{text\}\}'? & [Positivo, Negativo, Neutro, Misto] \\ 
p2 &  \mbox{task description} \mbox{+ question}  & Devi svolgere un compito di analisi del sentiment. Qual è il sentiment espresso nel seguente tweet: '\{\{text\}\}'? & [Positivo, Negativo, Neutro, Misto] \\ 
p3 & question \mbox{+ answer} & Qual è il sentiment espresso nel seguente tweet: '\{\{text\}\}'? 
   A: Positivo \textbackslash n B: Negativo \textbackslash n C: Neutro \textbackslash n D: Misto 
    \textbackslash n Risposta: & [A, B, C, D] \\ 
p4 & \mbox{task description} \mbox{+ question} \mbox{+ answer} & Devi svolgere un compito di analisi del sentiment. Qual è il sentiment espresso nel seguente tweet: '\{\{text\}\}'? A: Positivo \textbackslash n B: Negativo \textbackslash n C: Neutro \textbackslash n D: Misto \textbackslash n Risposta: & [A, B, C, D] \\ 
p5 &  affermative & Il seguente tweet: '\{\{text\}\}' esprime un sentiment & [Positivo, Negativo, Neutro, Misto]  \\ 
p6 & \mbox{task description} \mbox{+ affermative} & Devi svolgere un compito di analisi del sentiment. Il seguente tweet: '\{\{text\}\}' esprime un sentiment & [Positivo, Negativo, Neutro, Misto] \\ 
\botrule
\end{tabular}
\caption{Prompts for the Sentiment Analysis task.}
\label{tab:sentiment_prompts}
\end{table}

\subsection{Prompts for Generative Tasks}
A generative prompt requires the model to generate a portion of text as output, which is then evaluated with appropriate scoring metrics.
 For generative prompts, we adopted a compositional approach, combining three elements: (i) A request expressing the task to be executed by the model. This is a mandatory element for all generative prompts. (ii) A short description of the task. This element is optional and, if present, appears at the beginning of the prompt. (iii) Instructions for the output of the model. This is optional and, if included, appears at the end of the prompt.

 \begin{table}[ht]
\centering
\renewcommand{\arraystretch}{1.5} 
\begin{tabular}{c|p{2cm}|p{9cm}}
\toprule
\textbf{} & \textbf{Pattern} & \textbf{Prompt} \\ \midrule
p7 & request & Riassumi il seguente articolo di giornale:  '{{source}}' \textbackslash n Riassunto:  \\
p8 & request \mbox{+ output} & Estrai tutte le entità di tipo PER (persona), LOC (luogo) e ORG (organizzazione) dal testo seguente. Riporta ogni entità con il formato: Entità\$Tipo, separando ciascuna coppia con ','. Se non ci sono entità da estrarre, rispondi con '\&\&NOENT\&\&'. \textbackslash n Testo: '\{\{text\}\}' \textbackslash n Entità: \\ 
p9 & \mbox{task description} \mbox{+ request}  & Devi risolvere un compito di sintesi automatica del testo. Riassumi il seguente articolo di giornale:  '{{source}}' \textbackslash n Riassunto:  \\ 
p10 & \mbox{task description} \mbox{+ request} \mbox{+ output} & Devi svolgere un compito di riconoscimento delle entità nei testi. Estrai tutte le entità di tipo PER (persona), LOC (luogo) e ORG (organizzazione) dal testo seguente. Riporta ogni entità con il formato: Entità\$Tipo, separando ciascuna coppia con ','. Se non ci sono entità da estrarre, rispondi con '\&\&NOENT\&\&'. \textbackslash n Testo: '\{\{text\}\}' \textbackslash n Entità: \\
\botrule
\end{tabular}
\caption{Generative prompts used for the named entities recognition task (p8 and p10) and the summarization task (p7 and p9).}
\label{tab:entity_recognition_prompts}
\end{table}

Because of the high intensive computation required by generative tasks, we restricted the choice to four prompt templates, to be adapted to individual tasks.

\begin{itemize}
    \item \textit{Prompt 1: Request.} This prompt is formulated as a base request that the model has to perform in a generative mode. The request should respect the general guidelines for prompt definition (see Section \ref{sec:prompting}). As an example, the first prompt in Table \ref{tab:entity_recognition_prompts}, \textit{Riassumi il seguente articolo di giornale:  '{{source}}' \textbackslash n Riassunto:} is used for a summarization task.
    \item \textit{Prompt 2: Request + output.} This prompt is used when it is necessary to constrain the model's output to a precise format, in order to be properly parsed by the scoring metric. For instance, prompt p2 in Table \ref{tab:entity_recognition_prompts} requires that recognized named entities are presented in a list of pairs entity\$type, with each pair in turn separated by a ",".
    \item \textit{Prompt 3: Task description + request.} This prompt adds a short task description to a request prompt. For instance, prompt p3 in Table \ref{tab:entity_recognition_prompts} adds the description \textit{Devi risolvere un compito di sintesi automatica del testo.} to prompt p1.
    \item \textit{Prompt 4: Task description + request + output.} This is the most complete configuration for a generative prompt, where both the task description, the request and the instructions for out are present. As an example, this prompt has been used for the named entitie recognition task (see p4 in Table \ref{tab:entity_recognition_prompts}).
\end{itemize}

\subsection{Few-Shot Prompting}
Few-Shot prompting consists in including a small number of examples and their solution in the prompt given to the model. 
This approach has been widely adopted across various tasks using LLMs, as most of the times it improves performance compared to a Zero-Shot approach \cite{zhao2023survey}. 

Similarly to other aspects of prompting, several strategies have been proposed for incorporating examples into the prompt. 
One of the most popular methods is appending all examples at the end of the prompt.

\paragraph{Example:}

\textit{Devi svolgere un compito di riconoscimento delle entità nei testi. Estrai tutte le entità di tipo PER (persona), LOC (luogo) e ORG (organizzazione) dal testo seguente. Riporta ogni entità con il formato: Entità\$Tipo, separando ciascuna coppia con ','. Se non ci sono entità da estrarre, rispondi con '\&\&NOENT\&\&'. \\
Testo: L' ultimo affondo di quello che doveva essere un breve saluto finale e sulla scia delle polemiche diventa invece un vero discorso pubblico è per il Governo. \\
Entità: Governo\$ORG \\
Testo: — gridò ad Andrea.\\
Entità: Andrea\$PER}\\

On the other hand, the approach used by lm-evaluation-harness and others is to replicate the entire prompt for each few-shot example.
If we apply this format to the same example reported beforehand we obtain: 

\paragraph{Example:}

\textit{Devi svolgere un compito di riconoscimento delle entità nei testi. Estrai tutte le entità di tipo PER (persona), LOC (luogo) e ORG (organizzazione) dal testo seguente. Riporta ogni entità con il formato: Entità\$Tipo, separando ciascuna coppia con ','. Se non ci sono entità da estrarre, rispondi con '\&\&NOENT\&\&'. \\
Testo: L' ultimo affondo di quello che doveva essere un breve saluto finale e sulla scia delle polemiche diventa invece un vero discorso pubblico è per il Governo. \\
Entità: Governo\$ORG \\
Devi svolgere un compito di riconoscimento delle entità nei testi. Estrai tutte le entità di tipo PER (persona), LOC (luogo) e ORG (organizzazione) dal testo seguente. Riporta ogni entità con il formato: Entità\$Tipo, separando ciascuna coppia con '\%'. Se non ci sono entità da estrarre, rispondi con '\&\&NOENT\&\&'. \\
Testo: — gridò ad Andrea.\\
Entità: Andrea\$PER} \\

We noticed that the two few-shot modalities can affect the performance of the models. We experimented with both settings and we have not observed a clear indication of which format is better as results varied based on the task.
In the end, we decided to use the original implementation of Few-Shot prompting offered by LM-evaluation-harness. 

\section{Evaluation Metrics}\label{sec:metrics}

In this section, we describe the metrics that we used while developing the Evalita-LLM benchmark. According to \cite{mizrahi2024stateartmultipromptllm} we use three metrics to score multiple prompts given a task: maximum performance, average performance, and combined performance score.

We use the same notation proposed in \cite{mizrahi2024stateartmultipromptllm}: $M$ is a pretrained LLM, $T = \{(x_i , y_i )\}$ denotes an evaluation task  for the model $M$, $I_T$ is a set of prompts (i.e., instructions in natural language) for the task for $T$ , and $\epsilon(M,T,i) \in [0,1]$ denotes the aggregated performance of $M$ on samples from $T$, using a single instruction template $i \in I_T$ according to a standard metric, e.g., accuracy or $F_1$.

\paragraph{Minimum performance}
We define the minimum performance ($MinP$) of a certain prompt \(I\) in task \(T\) to be the minimum model performance achieved by the prompt \(I\) in all models.

\begin{equation}
    MinP_I(I,T,M_T) = \min_{m \in M_T}   \epsilon(I,T,m) 
    \label{eq:minperformance-prompt}
\end{equation}

\paragraph{Maximum performance}
We define the maximum performance ($MaxP$) (equation \ref{eq:maxperformance-llm}) of a model \(M\) on task \(T\) to be the maximum prompt performance that model \(M\) achieves across all prompt templates.

\begin{equation}
    MaxP_M(M,T,I_T) = \max_{i \in I_T}   \epsilon(M,T,i) 
    \label{eq:maxperformance-llm}
\end{equation}

\noindent
Equation \ref{eq:maxperformance-llm} can also be adapted to consider the maximum performance ($MaxP$) of a certain prompt \(I\) in task \(T\) to be the maximum model performance achieved by the prompt \(I\) in all models.

\begin{equation}
    MaxP_I(I,T,M_T) = \max_{m \in M_T}   \epsilon(I,T,m) 
    \label{eq:maxperformance-prompt}
\end{equation}

\noindent
This way, given a task task  \(T\), we can estimate both the best model on a set of different prompts, and the best prompt on a set of different models.

\paragraph{Average Performance}
We define the average performance (\(AvgP\)) (equation \ref{eq:averageperformance-llm}) of a model \(M\) on task \(T\) as the mean of the individual prompt performances over all prompt templates for the task \(T\):

\begin{equation}
    AvgP_M(M,T,IT) = \frac{1}{|I_T|} \cdot  \sum_{i \in I_T} \epsilon(M,T,i) 
    \label{eq:averageperformance-llm}
\end{equation}

\noindent
Similarly as equations \ref{eq:maxperformance-llm} and \ref{eq:maxperformance-prompt}, we adapt equation \ref{eq:averageperformance-llm} to estimate the average performance of a prompt \(I\)
on task \(T\) as the mean of the individual model performances over all models for the  task \(T\) (equation \ref{eq:averageperformance-prompt}):

\begin{equation}
    AvgP_I(I,T,M_T) = \frac{1}{|M_T|} \cdot  \sum_{m \in M_T} \epsilon(I,T,m) 
    \label{eq:averageperformance-prompt}
\end{equation}

\paragraph{Combined Performance Score}
In the same way that the $F_1$ score combines precision and recall into a single metric, we propose a Combined Performance Score (\(CPS\)) that unites the maximum and average performance metrics to capture both peak capability and robustness of the model across prompts. To define \(CPS\), we first introduce a model saturation score:

\begin{equation}
    Sat_M(M,T,IT) = 1 - (MaxP_M - AvgP_M) 
    \label{eq:saturation}
\end{equation}

\begin{equation}
    Sat_I(IT,T,M) = 1 - (MaxP_I - AvgP_I) 
    \label{eq:saturation}
\end{equation}

\noindent
This score measures how closely the model’s best performance aligns with its average performance. A high saturation score indicates that the model’s performance does not drop significantly for non-optimal instructions. Then, the CPS is calculated as the product of the model’s best performance (\(MaxP\)) and its saturation (\(Sat\)):

\begin{equation}
    CPS_M(M,T,IT) = Sat_M \cdot MaxP_M
    \label{eq:combinedperformance-llm}
\end{equation}

\noindent
Where equation \ref{eq:combinedperformance-prompt} is the version of equation \ref{eq:combinedperformance-llm} for prompt combination: 

\begin{equation}
    CPS_I(IT,T,M) = Sat_I \cdot MaxP_I
    \label{eq:combinedperformance-prompt}
\end{equation}

%
%

\section{Experiments}\label{sec:experiments}
This section reports on experiments conducted during the development of Evalita-LLMs. There are two main interrelated goals in the developing phase: 

\begin{itemize}
    \item \textit{Objective 1: validate a task against dev LLMs.} The intuition here is that a good task for the Evalita-LLM benchmark (and, in general, for any NLP benchmark), should be enough challenging for a set of selected LLMs, without being too difficult. This trade-off would ensure that the task would last enough time. To achieve this goal, we selected a set of LLMs for the dev phase and applied the metrics introduced in Section \ref{sec:metrics}.
    \item \textit{Objective 2: validate a set of prompts for a certain task.} On a certain degree, the performance of a LLM on a task depends on the prompts used. A relevant effort while developing Evalita-LLM was for prompt validation, i.e., ensuring that a certain set of prompt is correctly interpreted by an LLM, that different prompts produce enough different results to justify their presence in the benchmark, and that their execution time is compatible with reasonable computational resources. To achieve this goal, we used the six prompt templates described in Section \ref{sec:prompting}, and apply to them the metrics introduced in Section \ref{sec:metrics}.
\end{itemize}

\subsection{Large Language Models for Development (dev LLMs)}\label{sec:dev-llms}
To develop the Evalita-LLM benchmark, we run a number of experiments using few currently available LLMs. We selected six dev LLMs  with similar characteristics: they are open source and available on Hugging Face, all of them are in range 7B-9B, they are instructed versions to ensure reasonable interpretation of prompt instructions, and they have been pre-trained on some Italian data. As the goal of the experiments is not to rank the models for their performance, but rather to test and refine our methodological approach (i.e., selection of tasks and prompts), we keep the dev LLMs anonymized and use a placeholder for them (i.e., LLM-1, ... LLM-6).

\subsection{Experiments on Multiple-Choice Tasks}\label{sec:exp-mc}
Table \ref{tab:results-entailment} shows the results obtained during the dev phase on the Textual Entailment task. In order to validate the task (objective 1), we consider the performance of our six dev LLMs (see Section \ref{sec:dev-llms}). As for baselines for the task, we consider the random guess (50.00). All of the six dev LLMs outperform the two baselines (Average performance over the six prompts ranges from 53.50 to 70.08), with the Maximum performance (MaxP) as 78.75 for LLM-5. These data confirm that the task is well understood by the dev LLMs while still being challenging.

On the prompt side (top-right in Table \ref{tab:results-entailment}), we notice that accuracy scores range from a minimum of 45,25 for prompt p1 to a maximum of 78.75 for prompt p2  among the six models, while prompt p4 has both the highest average (63.54) and the highest CPS (66.37). In addition, there is high variability of scores within the same LLM: for instance, LLM-1 scores 55.00 with p1 and p2, and 70.25 with p3. Finally, the MaxP on single LLMs is achieved by 5 different prompts, out of the six we employed, showing that different models have different reactions to our prompts.

\begin{table}[ht]
\centering
\setlength{\tabcolsep}{3pt}  
\begin{tabular}{c|c|c|c|c|c|c||c|c|c|c}
\toprule
 & \textbf{LLM-1} & \textbf{LLM-2} & \textbf{LLM-3} & \textbf{LLM-4} & \textbf{LLM-5} & \textbf{LLM-6} & \textbf{MinP} & \textbf{MaxP} & \textbf{AvgP} & \textbf{CPS} \\ \midrule
\textbf{p1}   & 55.00 & 68.25 & 45.25 & 64.50 & 75.50 & 64.00 & \textbf{45.25} & 75.50 & 62.08 & 65.37 \\
\textbf{p2}   & 55.00 & 56.50 & 55.00 & 59.00 & 78.75 & 69.25 & 55.00 & \textbf{78.75} & 62.25 & 65.76 \\
\textbf{p3}   & 70.25 & 64.75 & 55.00 & 49.25 & 73.25 & 60.50 & 49.25 & 73.25 & 62.17 & 65.13 \\
\textbf{p4}   & 65.00 & 63.50 & 55.00 & 61.25 & 74.75 & 61.75 & 55.00 & 74.75 & \textbf{63.54} & \textbf{66.37} \\
\textbf{p5}   & 55.75 & 54.25 & 55.00 & 60.50 & 57.50 & 49.00 & 49.00 & 60.50 & 55.33 & 57.37 \\
\textbf{p6}   & 55.00 & 59.00 & 55.75 & 57.75 & 60.75 & 45.50 & 45.50 & 60.75 & 55.63 & 57.64 \\ 
\midrule
\textbf{MaxP} & 70.25 & 68.25 & 55.75 & 64.50 & \textbf{78.75} & 69.25 & \multicolumn{4}{c}{ }       \\
\textbf{AvgP} & 59.33 & 61.04 & 53.50 & 58.71 &\textbf{70.08} & 58.33 & \multicolumn{4}{c}{ }       \\
\textbf{CPS}  & 62.58 & 63.33 & 54.50 & 60.76 & \textbf{71.93} & 61.69 & \multicolumn{4}{c}{ }       \\ \botrule
\end{tabular}
\caption{Results during the dev phase: zero-shot $F_1$ on the Textual Entailment task.}
\label{tab:results-entailment}
\end{table}

As a second example for the multiple-choice tasks, Table \ref{tab:results-sentiment} shows the results obtained during the dev phase on the Sentiment Analysis task. As for baselines for the task, we consider random guess (25.00) and the most frequent (41.63). All of the six dev LLMs outperform the two baselines (Average performance over the six prompts ranges from 41.50 to 63.30), with the Maximum performance (MaxP) as 72.20 for LLM-6. These data confirm that the task is well understood by the dev LLMs while still being challenging.

On the prompt side (top-right in Table \ref{tab:results-sentiment}), we notice that $F_1$ scores range from a minimum of 26.30 for prompt p3 to a maximum of 72.20 for prompt p4  among the six models, while prompt p6 has both the highest average (63.27) and the highest CPS (64.91). In addition, there is high variability of scores within the same LLM: for instances LLM-4 scores 26.30 with p3, and 63.10 with p6. Finally, the MaxP on single LLMs is achieved by 3 different prompts (p4 for LLM-5 and LLM-6, p5 for LLM-2, and p6 for LLM-1, LLM-3 and LLM-4), out of the six we employed, showing that different models have different reactions to our prompt templates.

As a final comment, we notice that the MaxP and maximum AvgP values for the two tasks include three of our six prompt templates, showing that each prompt captures specific properties of the LLMs for the two tasks. We report all obtained results in the Appendix section.

\begin{table}[ht]
\centering
\setlength{\tabcolsep}{3pt}  
\begin{tabular}{c|c|c|c|c|c|c||c|c|c|c}
\toprule
 & \textbf{LLM-1} & \textbf{LLM-2} & \textbf{LLM-3} & \textbf{LLM-4} & \textbf{LLM-5} & \textbf{LLM-6} & \textbf{MinP} & \textbf{MaxP} & \textbf{AvgP} & \textbf{CPS} \\ \midrule
\textbf{p1}  & 33.06 & 43.90 & 36.93 & 52.99 & 41.14 & 41.19 & 33.06 & 52.99 & 41.54  & 46.92 \\
\textbf{p2}  & 37.19 & 49.35 & 36.71 & 38.69 & 59.40 & 49.82 & 36.71 & 59.40 & 45.19  & 50.96 \\
\textbf{p3}  & 53.33 & 64.18 & 43.24 & 26.27 & 65.64 & 69.72 & \textbf{26.27} & 69.72 & 53.73  & 58.57 \\
\textbf{p4}  & 54.35 & 68.31 & 46.74 & 28.31 & 70.73 & 72.17 & 28.31 & \textbf{72.17}  & 56.77  & 61.05 \\
\textbf{p5}  & 53.28 & 71.42 & 53.26 & 62.34 & 65.68 & 67.00 & 53.26 & 71.42 & 62.16  & 64.81 \\
\textbf{p6}  & 59.26 & 63.49 & 59.06 & 63.08 & 68.50 & 66.20 & 59.06 & 68.45 & \textbf{63.27} & \textbf{64.91} \\ \midrule
\textbf{MaxP} & 59,26 & 71,42 & 59,06 & 63,08 & 70,73 & \textbf{72,17} &  \multicolumn{4}{c}{ } \\
\textbf{AvgP} & 48,41 & 60,11 & 45,99 & 45,28 & \textbf{61,85} & 61,02 & \multicolumn{4}{c}{ } \\
\textbf{CPS}  & 52,83 & 63,34 & 51,34 & 51,85 & \textbf{64,45} & 64,12 &  \multicolumn{4}{c}{ }  \\ \botrule
\end{tabular}
\caption{Results during the dev phase: zero-shot $F_1$ on the Sentiment Analysis task.}
\label{tab:results-sentiment}
\end{table}

\subsection{Calculating Accuracy for Multiple-choice Tasks}

In this section we discuss how accuracy is calculated for multiple-choice tasks, highlighting relevant featured implemented in the LLM-evaluation-harness library. Given a multiple-choice question, the objective is to identify the most probable correct answer. The steps involved are yje following:

\begin{enumerate}
    \item Combine the question (input \( x \)) and each answer option (\( y \)).
    \item Find the chance that each answer is correct using:
    \[
    \log P(y \mid x) = \sum_{i=0}^{m-1} \log P(y_i \mid x, y_0, \dots, y_{i-1})
    \]
    where \( m \) is the number of tokens (or words) in the answer \( y \). This formula calculates the log probability for each token in the answer, given the context provided by the question and the preceding tokens in the answer.
    \item Pick the answer with the highest probability:
    \[
    \text{Best Answer} = \arg\max \big( \log P(a_1 \mid x), \log P(a_2 \mid x), \dots, \log P(a_k \mid x) \big)
    \]
\end{enumerate}

This method, referred to as \textit{acc} in lm-evaluation-harness, is used for all multiple-choice tasks. However, it can penalize longer responses. This happens because the model calculates the score by summing the probabilities for each word. Since the logarithm of a probability is always negative (or zero for a probability of 1), summing multiple negative logarithms results in a more negative total as the answer length increases. This leads to a lower overall score for longer answers. To address this issue, a normalization can be applied to adjust the scores. One commonly used normalization approach is \textit{Byte-length Normalization}, implemented in lm-evaluation-harness for all multiple-choice tasks and referred to as \textit{acc\_norm}. This approach divides the score by the size of the answer in bytes. In practice, for most English texts (which typically use 1 byte per character in UTF-8), dividing by the number of bytes is very similar to dividing by the number of characters.

\[
\text{Byte-length Normalized Score} = \frac{\log P(y \mid x)}{\text{Answer Size in Bytes}}
\] \\

The Evalita-LLM benchmark includes three tasks evaluated with the accuracy measure: Textual Entailment (TE), Admission Tests (AT) and Adaptation of Frequently Asked Questions (FAQ). Two of these tasks (AT and FAQ) involve a mix of short and long answers, leading to a significant difference when the results are calculated using \textit{acc} instead of \textit{acc\_norm}. To highlight these differences, Tables \ref{tab:acc_vs_acc_norm_at} and \ref{tab:acc_vs_acc_norm_faq} present the results obtained for the AT task and FAQ task using both \textit{acc} and \textit{acc\_norm}. In the tables, for prompts p3 and p4, the answers are simple letter choices (A, B, C, D or E). These are short and typically consist of a single character, meaning there is minimal byte-length for the normalization to adjust. For prompts p1, p2, p5 and p6, the answers consist of both the letter choice and the full text of the corresponding options, which can vary in length for the same question (e.g., A: Place a chest drain in the 2nd intercostal space at the anterior axillary line, B: Intubate the patient). This increases the byte-length of the answers significantly. Table \ref{tab:acc_vs_acc_norm_te} reports the results for the TE task. For prompts p1, p2, p3, and p4, where the answers have comparable lengths, there are no significant differences between the accuracy values (\textit{acc}) and normalized accuracy (\textit{acc\_norm}). However, for prompts p5 and p6, where slight variations in answer length are present, the results obtained with \textit{acc\_norm} are unexpectedly lower than those obtained with \textit{acc} for most models.

\begin{table}[ht]
\centering
\setlength{\tabcolsep}{3pt}  
\begin{tabular}{c|c|c|c|c|c|c}
\toprule
 & \textbf{LLM-1} & \textbf{LLM-2} & \textbf{LLM-3} & \textbf{LLM-4} & \textbf{LLM-5} & \textbf{LLM-6} \\ \midrule
\textbf{p1} & 22.00 (28.40) & 24.40 (29.80) & 22.04 (26.25) & 26.20 (36.20) & 39.00 (43.60) & 27.40 (31.00) \\
\textbf{p2} & 23.20 (28.20) & 26.00 (28.80) & 20.84 (25.25) & 24.20 (31.40) & 37.40 (42.80) & 26.80 (30.60) \\
\textbf{p3} & 31.20 (31.20) & 67.00 (67.00) & 29.06 (29.06) & 53.00 (53.00) & 77.70 (77.70) & 62.20 (62.20) \\
\textbf{p4} & 30.40 (30.40) & 68.60 (68.60) & 24.85 (24.85) & 51.40 (51.40) & 77.20 (77.20) & 62.20 (62.20) \\
\textbf{p5} & 25.80 (31.40) & 32.40 (37.60) & 23.60 (30.40) & 34.40 (37.20) & 39.80 (44.00) & 33.40 (37.80) \\
\textbf{p6} & 24.20 (30.40) & 33.00 (35.40) & 24.40 (29.40) & 34.60 (37.60) & 38.80 (45.00) & 32.40 (37.40) \\
\bottomrule
\end{tabular}
\caption{Comparison of acc and acc\_norm (in parentheses) on the AT task in the zero-shot experiment.}
\label{tab:acc_vs_acc_norm_at}
\end{table}

\begin{table}[ht]
\centering
\setlength{\tabcolsep}{3pt}  
\begin{tabular}{c|c|c|c|c|c|c}
\toprule
 & \textbf{LLM-1} & \textbf{LLM-2} & \textbf{LLM-3} & \textbf{LLM-4} & \textbf{LLM-5} & \textbf{LLM-6} \\ \midrule
\textbf{p1} & 28.43 (35.41) & 30.42 (46.38) & 28.93 (44.14) & 29.43 (44.39) & 28.93 (39.90) & 30.42 (47.13) \\
\textbf{p2} & 28.18 (35.66) & 30.42 (44.89) & 28.68 (44.89) & 28.93 (44.64) & 28.18 (39.15) & 30.92 (45.89) \\
\textbf{p3} & 35.91 (35.91) & 80.30 (80.30) & 27.43 (27.43) & 31.92 (31.92) & 91.27 (91.27) & 59.60 (59.60) \\
\textbf{p4} & 38.90 (38.90) & 87.78 (87.78) & 29.43 (29.43) & 57.36 (57.36) & 98.00 (98.00) & 71.82 (71.82) \\
\textbf{p5} & 28.18 (36.41) & 30.92 (46.63) & 29.43 (46.13) & 29.68 (45.89) & 28.33 (38.15) & 31.17 (49.38) \\
\textbf{p6} & 27.93 (34.91) & 31.17 (45.89) & 28.93 (45.14) & 29.43 (44.64) & 28.68 (39.90) & 32.92 (51.87) \\
\bottomrule
\end{tabular}
\caption{Comparison of acc and acc\_norm (in parentheses) on the FAQ task in the zero-shot experiment.}
\label{tab:acc_vs_acc_norm_faq}
\end{table}

\begin{table}[ht]
\centering
\setlength{\tabcolsep}{3pt}  
\begin{tabular}{c|c|c|c|c|c|c}
\toprule
 & \textbf{LLM-1} & \textbf{LLM-2} & \textbf{LLM-3} & \textbf{LLM-4} & \textbf{LLM-5} & \textbf{LLM-6} \\ \midrule
\textbf{p1} & 55.00 (55.00) & 68.25 (68.25) & 45.25 (45.25) & 64.50 (64.50) & 75.50 (75.50) & 64.00 (64.00) \\ 
\textbf{p2} & 55.00 (55.00) & 56.50 (56.50) & 55.00 (55.00) & 59.00 (59.00) & 78.75 (78.75) & 69.25 (69.25) \\ 
\textbf{p3} & 70.25 (70.25) & 64.75 (64.75) & 55.00 (55.00) & 49.25 (49.25) & 73.25 (73.25) & 60.50 (60.50) \\ 
\textbf{p4} & 65.00 (65.00) & 63.50 (63.50) & 55.00 (55.00) & 61.25 (61.25) & 74.75 (74.75) & 61.75 (61.75) \\ 
\textbf{p5} & 55.75 (59.00) & 54.25 (45.50) & 55.00 (49.50) & 60.50 (50.50) & 57.50 (47.50) & 49.00 (45.50) \\ 
\textbf{p6} & 55.00 (55.00) & 59.00 (50.50) & 55.75 (50.00) & 57.75 (51.50) & 60.75 (53.25) & 45.50 (45.50) \\ 
\bottomrule
\end{tabular}
\caption{Comparison of acc and acc\_norm (in parentheses) on the TE task in the zero-shot experiment}
\label{tab:acc_vs_acc_norm_te}
\end{table}

\subsection{Experiments on Generative Tasks}\label{sec:exp-gen}
Table \ref{tab:results-NER} shows the results obtained during the dev phase on the Named Entity Recognition task, which we implemented as a generative task. The average $F_1$ performance, in the zero-shot configuration, over the six LLMs used for dev, ranges between 14.03 (LLM-1) and 39.78 (LLM-4), with a maximum performance of 48.31 (LLM-6). Although performance is significantly lower than fine-tuned models used as baselines for the task (i.e., CRF with gazetteers achieves 0.83, while fine-tuned BERT is 0.89), we believe that the zero-shot generative setting in our benchmark is both interesting and challenging for large pre-trained LLMs.

As far as prompts are concerned, p8 (request + output, see Section \ref{sec:prompting}) performs better than p10 (task description + request + output) in all LLMs but LLM-4, achieving higher MaxP, AvgP and CPS. On the other hand, p10 gets the minimum performance on LLM-1.

Finally, as general comment on generative tasks, forcing an LLM to a certain output is more complex than expected. For instance, conditioning the output for the NER task, i.e., each entity has to be reported as a pair entity\$type, and each pair has to be separated by a comma, like [Bill Clinton\$PER, USA\$ORG], may work relatively well for a certain model, and very bad for another model, independently of the capacity of the two models to recognize persons and organizations.

\begin{table}[ht]
\centering
\setlength{\tabcolsep}{3pt}  
\begin{tabular}{c|c|c|c|c|c|c||c|c|c|c}
\toprule
 & \textbf{LLM-1} & \textbf{LLM-2} & \textbf{LLM-3} & \textbf{LLM-4} & \textbf{LLM-5} & \textbf{LLM-6} & \textbf{MinP} & \textbf{MaxP} & \textbf{AvgP} & \textbf{CPS} \\ \midrule
\textbf{p8}     & 20.61          & 15.51          & 23.98          & 39.49          & 40.46          & 48.31          & 15.51 & \textbf{48.31} & \textbf{31.39} &  \textbf{40.14}\\
\textbf{p10}     & 7.45           & 13.12          & 28.51          & 40.07          & 34.91          & 28.20          & \textbf{7.45}  & 40.07 & 25.38 & 34.18  \\ \midrule
\textbf{MaxP}   & 20.61          & 15.51          & 28.51          & 40.07          & 40.46          & 48.31      &    \multicolumn{4}{c}{ }                \\
\textbf{AvgP}   & 14.03          & 14.32          & 26.25          & 39.78          & 37.69          & 38.26    &      \multicolumn{4}{c}{ }                \\
\textbf{CPS}    & 19.25          & 15.32          & 27.86          & 39.95          & 39.34          & 43.45    &     \multicolumn{4}{c}{ }                \\ \bottomrule
\end{tabular}
\caption{Results during the dev phase: zero-shot $F_1$ on the Named Entities Recognition task.}
\label{tab:results-NER}
\end{table}

\subsection{Discussion}\label{sec:discussion}
Tables \ref{tab:mc-taskdistribution} and  \ref{tab:gen-taskdistribution} show the distribution of the multiple-choice tasks and of the generative tasks, over the respective multiple-choice prompts and the four metrics.

As for multiple-choice tasks, we see that p1 shows the majority of minimum performances across three tasks: Hate Speech (HS), Textual Entailment (TE) and Word in Context (WIC). p2 demonstrates both a minimum performance on the Admission Test (AT) and a maximum performance on TE. p3 has minimum performances on two tasks, Sentiment Analysis (SA) and Frequently Asked Questions (FAQ), as well as the best average and CSP for AT. p4 has the highest density of MaxP (four tasks), the highest average and CSP across two tasks. p5 achieves both the best average and CSP for HS. Finally, p6 performs the best on WIC, and gets the highest average and CPS for SA. We also notice that each of the four metrics we implemented is covered by at least three prompt templates. Overall, the distribution reported in Table \ref{tab:mc-taskdistribution} shows that all prompt templates capture different characteristics of the six tasks, empirically confirming our prompt selection.

As far as generative tasks, Table \ref{tab:gen-taskdistribution} shows the performance distribution of the four generative tasks over the four generative prompts we used. p7 (request only), as expected, does not reach any MaxP on our tasks, because it does not specify any output format. p8 (request + output format) is notable because it achieves both the minimum and the maximum performance on the same two tasks: Lexical Substitution (LS) and Relation Extraction (REL). In addition, it achieves the MaxP, best AvgP and best CPS on the Named Entity Recognition (NER) task, showing that for NER and REL the output format is critical. p9 (task description + request) is the best prompt template for Summarization (SUM), showing that the task does not benefit from output definition. This is likely due to the fact that our dev LLMs were already instructed to produce summaries. p10 (task description + request + output format) is at the same time the worst prompt template for NER and the best average and CPS for Lexical Substitution (LS).

\begin{table}[ht]
\centering
\caption{Multiple-choice task distribution over the six prompts and four metrics.}
\label{tab:mc-taskdistribution}
\begin{tabular}{c|c|c|c|c}
\toprule
\textbf{} & \textbf{MinP} & \textbf{MaxP} & \textbf{AvgP} & \textbf{CPS} \\ \midrule
\textbf{p1} & HS TE WIC & ---    & ---        & ---        \\
\textbf{p2} & AT        & TE     & ---        & ---        \\
\textbf{p3} & SA FAQ    & ---    & AT         & AT         \\
\textbf{p4} & ---       & SA HS AT FAQ & TE FAQ & TE FAQ \\
\textbf{p5} & ---       & ---    & HS         & HS         \\
\textbf{p6} & ---       & WIC    & SA WIC     & SA WIC     \\ \bottomrule
\end{tabular}
\end{table}

\begin{table}[ht]
\centering
\caption{Generative task distribution over the four prompts and four metrics.}
\label{tab:gen-taskdistribution}
\begin{tabular}{c|c|c|c|c}
\toprule
\textbf{} & \textbf{MinP}   & \textbf{MaxP}       & \textbf{AvgP}   & \textbf{CPS}   \\ \midrule
\textbf{p7} & SUM             & ---                 & ---             & ---            \\
\textbf{p8} & LS REL          & LS NER REL          & NER             & NER REL        \\
\textbf{p9} & ---             & SUM                 & SUM             & SUM            \\
\textbf{p10} & NER            & ---                 & LS REL          & LS             \\ \bottomrule
\end{tabular}
\end{table}

\subsection{Running Time}
In addition to assessing the utility of a set of prompts, we are also interested in ensuring that the time required to run the whole benchmark on a certain model is reasonable.
Table \ref{tab:running-time} shows the running time for all tasks (six prompts for the multiple-choice tasks and two prompts for the generative tasks) in the Evalita-LLM benchmark.

\begin{table}[h]
    \centering
    \begin{tabular}{c|c|c|c|c|c|c}
        \hline
        & \textbf{LLM-1} & \textbf{LLM-2} & \textbf{LLM-3} & \textbf{LLM-4 }& \textbf{LLM-5} & \textbf{LLM-6} \\
        \hline
         batch-size & 1   &  4    & 8    & 8    & 2    & 4 \\
         time (hours)      & 13.6 &  29.3 & 15.1 & 15.7 & 22.8 & 24.0 \\
    \bottomrule
    \end{tabular}
    \caption{Running time (in hours) for all tasks in the benchmark on a single NVIDIA A40 GPU, performed using zero-shot evaluation. The batch size used for evaluation is also shown and was chosen to ensure the model fits within the GPU memory.}
    \label{tab:running-time}
\end{table}

\section{Conclusion}
\label{sec:conclusion}
We described Evalita-LLM, a new benchmark designed to evaluate Large Language Models on Italian tasks. The distinguishing, and innovative, features of Evalita-LLM are the following: (i) all tasks are native Italian, avoiding issues of translating from Italian and potential cultural biases; (ii) in addition to multiple-choice tasks,  we include generative tasks, aiming at a more natural interaction with the model; (iii) all tasks are evaluated against multiple prompts, this way mitigating the model sensitivity to similar prompts and allowing a more fair and objective evaluation. 
\\
\bmhead{Acknowledgements}
This work has been partially supported by the PNRR project FAIR - Future AI Research (PE00000013), under the NRRP MUR
program funded by NextGenerationEU.


\begin{appendices}
\section{Details of dev experiments for the ten Evalita-LLM tasks}
\label{sec:A1}

\begin{table}[ht]
\centering
\setlength{\tabcolsep}{4pt}  
\begin{tabular}{c|c c|c c|c c|c c|c c|c c}
\toprule
 & \multicolumn{2}{c|}{\textbf{LLM 1}} & \multicolumn{2}{c|}{\textbf{LLM 2}} & \multicolumn{2}{c|}{\textbf{LLM 3}} & \multicolumn{2}{c|}{\textbf{LLM 4}} & \multicolumn{2}{c|}{\textbf{LLM 5}} & \multicolumn{2}{c}{\textbf{LLM 6}} \\ \midrule
 & \textit{ZS} & \textit{FS} & \textit{ZS} & \textit{FS} & \textit{ZS} & \textit{FS} & \textit{ZS} & \textit{FS} & \textit{ZS} & \textit{FS} & \textit{ZS} & \textit{FS} \\ \midrule
\textbf{P1}  & 55.00 & 57.25 & 68.25 & 73.75 & 45.25 & 52.00 & 64.50 & 73.75 & 75.50 & 80.75 & 64.00 & 76.75 \\
\textbf{P2}  & 55.00 & 57.50 & 56.50 & 74.00 & 55.00 & 52.50 & 59.00 & 74.25 & 78.75 & 81.50 & 69.25 & 74.25 \\
\textbf{P3}  & 70.25 & 58.50 & 64.75 & 73.00 & 55.00 & 55.00 & 49.25 & 68.25 & 73.25 & 80.25 & 60.50 & 72.00 \\
\textbf{P4}  & 65.00 & 61.25 & 63.50 & 75.25 & 55.00 & 54.50 & 61.25 & 66.25 & 74.75 & 81.75 & 61.75 & 68.50 \\
\textbf{P5}  & 55.75 & 51.25 & 54.25 & 60.00 & 55.00 & 51.50 & 60.50 & 66.00 & 57.50 & 75.75 & 49.00 & 68.75 \\
\textbf{P6}  & 55.00 & 49.25 & 59.00 & 57.75 & 55.75 & 53.75 & 57.75 & 67.75 & 60.75 & 76.25 & 45.50 & 69.00 \\ 
\midrule
\textbf{MaxP} & 70.25 & 61.25 & 68.25 & 75.25 & 55.75 & 55.00 & 64.50 & 74.25 & 78.75 & \textbf{81.75} & 69.25 & 76.75 \\
\textbf{AvgP} & 59.33 & 55.83 & 61.04 & 68.96 & 53.50 & 53.21 & 58.71 & 69.38 & 70.08 & \textbf{79.38} & 58.33 & 71.54 \\
\textbf{CPS}  & 62.58 & 57.93 & 63.33 & 70.52 & 54.50 & 54.01 & 60.76 & 70.63 & 71.93 & \textbf{79.81} & 61.69 & 72.75 \\
\bottomrule
\end{tabular}
\caption{LLM Results with Zero-shot (ZS) and Few-shot (FS) Evaluations for the Textual Entailment task.}
\label{tab:te_results}
\end{table}

\begin{table}[ht]
\centering
\setlength{\tabcolsep}{4pt}  
\begin{tabular}{c|c c|c c|c c|c c|c c|c c}
\toprule
 & \multicolumn{2}{c|}{\textbf{LLM 1}} & \multicolumn{2}{c|}{\textbf{LLM 2}} & \multicolumn{2}{c|}{\textbf{LLM 3}} & \multicolumn{2}{c|}{\textbf{LLM 4}} & \multicolumn{2}{c|}{\textbf{LLM 5}} & \multicolumn{2}{c}{\textbf{LLM 6}} \\ \midrule
 & \textit{ZS} & \textit{FS} & \textit{ZS} & \textit{FS} & \textit{ZS} & \textit{FS} & \textit{ZS} & \textit{FS} & \textit{ZS} & \textit{FS} & \textit{ZS} & \textit{FS} \\ \midrule
\textbf{P1}  & 33.06 & 66.96 & 43.90 & 69.38 & 36.93 & 56.74 & 52.99 & 69.07 & 41.14 & 73.96 & 41.19 & 68.19 \\
\textbf{P2}  & 37.19 & 67.21 & 49.35 & 70.19 & 36.71 & 52.84 & 38.69 & 68.24 & 59.40 & 73.82 & 49.82 & 69.64 \\
\textbf{P3}  & 53.33 & 72.52 & 64.18 & 77.39 & 43.24 & 39.76 & 26.27 & 67.84 & 65.64 & 71.32 & 69.72 & 74.77 \\
\textbf{P4}  & 54.35 & 71.48 & 68.31 & 77.39 & 46.74 & 45.13 & 28.31 & 67.96 & 70.73 & 71.62 & 72.17 & 75.96 \\
\textbf{P5}  & 53.28 & 68.97 & 71.42 & 70.38 & 53.26 & 62.88 & 62.34 & 69.63 & 65.68 & 72.25 & 67.00 & 66.37 \\
\textbf{P6}  & 59.26 & 70.34 & 63.49 & 72.84 & 59.06 & 59.70 & 63.08 & 72.04 & 68.50 & 72.56 & 66.20 & 70.14 \\ 
\midrule
\textbf{MaxP} & 59.26 & 72.52 & 71.42 & \textbf{77.39} & 59.06 & 62.88 & 63.08 & 72.04 & 70.73 & 73.96 & 72.17 & 75.96 \\
\textbf{AvgP} & 48.41 & 69.58 & 60.11 & 72.93 & 45.99 & 52.84 & 45.28 & 69.13 & 61.85 & \textbf{72.59} & 61.02 & 70.85 \\
\textbf{CPS}  & 52.83 & 70.39 & 63.34 & \textbf{73.94} & 51.34 & 56.57 & 51.85 & 69.94 & 64.45 & 72.95 & 64.12 & 72.07 \\
\bottomrule
\end{tabular}
\caption{LLM Results with Zero-shot (ZS) and Few-shot (FS) Evaluations for the Sentiment Analysis Task.}
\label{tab:sa_results}
\end{table}

\begin{table}[ht]
\centering
\setlength{\tabcolsep}{4pt}  
\begin{tabular}{c|c c|c c|c c|c c|c c|c c}
\toprule
 & \multicolumn{2}{c|}{\textbf{LLM 1}} & \multicolumn{2}{c|}{\textbf{LLM 2}} & \multicolumn{2}{c|}{\textbf{LLM 3}} & \multicolumn{2}{c|}{\textbf{LLM 4}} & \multicolumn{2}{c|}{\textbf{LLM 5}} & \multicolumn{2}{c}{\textbf{LLM 6}} \\ \midrule
 & \textit{ZS} & \textit{FS} & \textit{ZS} & \textit{FS} & \textit{ZS} & \textit{FS} & \textit{ZS} & \textit{FS} & \textit{ZS} & \textit{FS} & \textit{ZS} & \textit{FS} \\ \midrule
\textbf{P1}  & 60.85 & 41.33 & 15.13 & 52.79 & 59.55 & 29.25 & 61.45 & 48.84 & 58.40 & 71.31 & 0.25 & 67.00 \\
\textbf{P2}  & 48.78 & 40.47 & 40.60 & 58.57 & 32.45 & 37.46 & 61.79 & 53.30 & 49.73 & 70.48 & 2.43 & 69.47 \\
\textbf{P3}  & 55.05 & 61.90 & 36.68 & 69.96 & 61.00 & 61.30 & 1.94 & 65.75 & 68.81 & 76.20 & 42.76 & 66.31 \\
\textbf{P4}  & 62.68 & 63.20 & 52.66 & 72.22 & 61.77 & 60.60 & 49.84 & 66.28 & 69.39 & 76.01 & 62.47 & 74.22 \\
\textbf{P5}  & 63.58 & 34.90 & 63.17 & 65.97 & 62.85 & 44.04 & 64.24 & 54.90 & 63.39 & 65.92 & 63.39 & 71.07 \\
\textbf{P6}  & 62.43 & 39.21 & 62.59 & 64.43 & 44.43 & 46.29 & 62.96 & 54.43 & 62.44 & 64.97 & 66.18 & 71.08 \\
\midrule
\textbf{MaxP} & 63.58 & 63.20 & 63.17 & 72.22 & 62.85 & 61.30 & 64.24 & 66.28 & 69.39 & \textbf{76.20} & 66.18 & 74.22 \\
\textbf{AvgP} & 58.90 & 46.84 & 45.14 & 63.99 & 53.68 & 46.49 & 50.37 & 57.25 & 62.03 & \textbf{70.82} & 39.58 & 69.86 \\
\textbf{CPS}  & 60.60 & 52.86 & 51.78 & 66.28 & 57.08 & 52.22 & 55.33 & 60.29 & 64.28 & \textbf{72.10} & 48.58 & 70.98 \\
\bottomrule
\end{tabular}
\caption{LLM Results with Zero-shot (ZS) and Few-shot (FS) Evaluations for the Hate Speech Task.}
\label{tab:hs_results}
\end{table}

\begin{table}[ht]
\centering
\setlength{\tabcolsep}{4pt}  
\begin{tabular}{c|c c|c c|c c|c c|c c|c c}
\toprule
 & \multicolumn{2}{c|}{\textbf{LLM 1}} & \multicolumn{2}{c|}{\textbf{LLM 2}} & \multicolumn{2}{c|}{\textbf{LLM 3}} & \multicolumn{2}{c|}{\textbf{LLM 4}} & \multicolumn{2}{c|}{\textbf{LLM 5}} & \multicolumn{2}{c}{\textbf{LLM 6}} \\ \midrule
 & \textit{ZS} & \textit{FS} & \textit{ZS} & \textit{FS} & \textit{ZS} & \textit{FS} & \textit{ZS} & \textit{FS} & \textit{ZS} & \textit{FS} & \textit{ZS} & \textit{FS} \\ \midrule
\textbf{P1}  & 22.00 & 28.20 & 24.40 & 35.20 & 22.04 & 28.00 & 26.20 & 36.20 & 39.00 & 44.20 & 27.40 & 37.20 \\
\textbf{P2}  & 23.20 & 28.40 & 26.00 & 35.60 & 20.84 & 27.00 & 24.20 & 35.80 & 37.40 & 44.40 & 26.80 & 37.60 \\
\textbf{P3}  & 31.20 & 38.00 & 67.00 & 70.60 & 29.06 & 32.00 & 53.00 & 57.00 & 77.70 & 75.60 & 62.20 & 64.80 \\
\textbf{P4}  & 30.40 & 36.20 & 68.60 & 69.60 & 24.85 & 32.60 & 51.40 & 55.80 & 77.20 & 75.60 & 62.20 & 64.40 \\
\textbf{P5}  & 25.80 & 30.00 & 32.40 & 36.00 & 23.60 & 28.80 & 34.40 & 38.20 & 39.80 & 44.80 & 33.40 & 37.60 \\
\textbf{P6}  & 24.20 & 29.20 & 33.00 & 35.60 & 24.40 & 28.20 & 34.60 & 36.80 & 38.80 & 43.80 & 32.40 & 38.20 \\
\midrule
\textbf{MaxP} & 31.20 & 38.00 & 68.60 & 70.60 & 29.06 & 32.60 & 53.00 & 57.00 & 77.70 & 75.60 & 62.20 & 64.80 \\
\textbf{AvgP} & 26.13 & 31.67 & 41.90 & 47.10 & 24.13 & 29.43 & 37.30 & 43.30 & 51.65 & 54.73 & 40.73 & 46.63 \\
\textbf{CPS} & 29.62 & 35.59 & 50.28 & 54.01 & 27.63 & 31.57 & 44.68 & 49.19 & 57.46 & 59.82 & 48.85 & 53.03 \\
\bottomrule
\end{tabular}
\caption{LLM Results with Zero-shot (ZS) and Few-shot (FS) Evaluations for the Admission Test Task.}
\label{tab:at_results}
\end{table}

\begin{table}[ht]
\centering
\setlength{\tabcolsep}{4pt}  
\begin{tabular}{c|c c|c c|c c|c c|c c|c c}
\toprule
 & \multicolumn{2}{c|}{\textbf{LLM 1}} & \multicolumn{2}{c|}{\textbf{LLM 2}} & \multicolumn{2}{c|}{\textbf{LLM 3}} & \multicolumn{2}{c|}{\textbf{LLM 4}} & \multicolumn{2}{c|}{\textbf{LLM 5}} & \multicolumn{2}{c}{\textbf{LLM 6}} \\ \midrule
 & \textit{ZS} & \textit{FS} & \textit{ZS} & \textit{FS} & \textit{ZS} & \textit{FS} & \textit{ZS} & \textit{FS} & \textit{ZS} & \textit{FS} & \textit{ZS} & \textit{FS} \\ \midrule
\textbf{P1}  & 66.67 & 47.08 & 55.18 & 59.54 & 0.00 & 39.41 & 62.78 & 31.55 & 3.92 & 55.74 & 53.99 & 64.52 \\
\textbf{P2}  & 66.67 & 42.82 & 53.23 & 60.67 & 42.79 & 24.85 & 30.85 & 22.37 & 9.16 & 50.79 & 12.23 & 57.78 \\
\textbf{P3}  & 7.43  & 53.93 & 54.55 & 58.44 & 64.47 & 61.81 & 32.86 & 39.90 & 40.12 & 61.90 & 64.36 & 50.69 \\
\textbf{P4}  & 26.22 & 52.53 & 6.87  & 58.91 & 65.67 & 58.15 & 63.23 & 49.07 & 2.37  & 48.75 & 57.97 & 33.93 \\
\textbf{P5}  & 66.13 & 18.95 & 65.39 & 52.84 & 66.31 & 48.78 & 66.76 & 20.67 & 64.86 & 59.25 & 59.75 & 52.81 \\
\textbf{P6}  & 66.40 & 40.69 & 66.93 & 56.35 & 65.55 & 37.50 & 64.43 & 20.74 & 62.10 & 59.74 & 66.57 & 58.45 \\
\midrule
\textbf{MaxP} & 66.67 & 53.93 & \textbf{66.93} & 60.67 & 66.31 & 61.81 & 66.76 & 49.07 & 64.86 & 61.90 & 66.57 & 64.52 \\
\textbf{AvgP} & 49.92 & 42.67 & 50.36 & \textbf{57.79} & 50.80 & 45.08 & 53.49 & 30.72 & 30.42 & 56.03 & 52.48 & 53.03 \\
\textbf{CPS} & 55.50 & 47.86 & 55.84 & 58.92 & 56.02 & 51.47 & 57.90 & 40.06 & 42.52 & \textbf{58.27} & 57.19 & 57.11 \\
\bottomrule
\end{tabular}
\caption{LLM Results with Zero-shot (ZS) and Few-shot (FS) Evaluations for the Word in Context Task.}
\label{tab:wic_results}
\end{table}

\begin{table}[ht]
\centering
\setlength{\tabcolsep}{4pt}  
\begin{tabular}{c|c c|c c|c c|c c|c c|c c}
\toprule
 & \multicolumn{2}{c|}{\textbf{LLM 1}} & \multicolumn{2}{c|}{\textbf{LLM 2}} & \multicolumn{2}{c|}{\textbf{LLM 3}} & \multicolumn{2}{c|}{\textbf{LLM 4}} & \multicolumn{2}{c|}{\textbf{LLM 5}} & \multicolumn{2}{c}{\textbf{LLM 6}} \\ \midrule
 & \textit{ZS} & \textit{FS} & \textit{ZS} & \textit{FS} & \textit{ZS} & \textit{FS} & \textit{ZS} & \textit{FS} & \textit{ZS} & \textit{FS} & \textit{ZS} & \textit{FS} \\ \midrule
\textbf{P1}  & 28.43 & 28.68 & 30.42 & 29.68 & 28.93 & 29.18 & 29.43 & 28.93 & 28.93 & 27.43 & 30.42 & 34.16 \\
\textbf{P2}  & 28.18 & 28.43 & 30.42 & 29.68 & 28.68 & 29.43 & 28.93 & 28.68 & 28.18 & 27.43 & 30.92 & 35.41 \\
\textbf{P3}  & 35.91 & 41.50 & 80.30 & 94.00 & 27.43 & 30.92 & 31.92 & 40.15 & 91.27 & 98.75 & 59.60 & 88.78 \\
\textbf{P4}  & 38.90 & 50.12 & 87.78 & 93.52 & 29.43 & 36.16 & 57.36 & 52.12 & 98.00 & 98.00 & 71.82 & 88.28 \\
\textbf{P5}  & 28.18 & 28.43 & 30.92 & 29.93 & 29.43 & 28.43 & 29.68 & 28.93 & 28.33 & 27.43 & 31.17 & 32.92 \\
\textbf{P6}  & 27.93 & 28.43 & 31.17 & 29.43 & 28.93 & 29.43 & 29.43 & 28.93 & 28.68 & 27.68 & 32.92 & 35.16 \\
\midrule
\textbf{MaxP} & 38.90 & 50.12 & 87.78 & 93.52 & 29.43 & 36.16 & 57.36 & 52.12 & 98.00 & \textbf{98.75} & 71.82 & 88.78 \\
\textbf{AvgP} & 31.26 & 34.27 & 48.50 & 36.03 & 28.81 & 30.59 & 34.46 & 34.62 & 50.57 & 51.12 & 42.81 & \textbf{52.45} \\
\textbf{CPS} & 35.93 & 42.17 & 53.30 & 39.75 & 29.25 & 34.15 & 44.22 & 43.00 & 51.51 & 51.72 & 50.98 & \textbf{56.53} \\
\bottomrule
\end{tabular}
\caption{LLM Results with Zero-shot (ZS) and Few-shot (FS) Evaluations for the FAQ Task.}
\label{tab:faq_results}
\end{table}

\begin{table}[ht]
\centering
\setlength{\tabcolsep}{4pt}  
\begin{tabular}{c|c c|c c|c c|c c|c c|c c}
\toprule
 & \multicolumn{2}{c|}{\textbf{LLM 1}} & \multicolumn{2}{c|}{\textbf{LLM 2}} & \multicolumn{2}{c|}{\textbf{LLM 3}} & \multicolumn{2}{c|}{\textbf{LLM 4}} & \multicolumn{2}{c|}{\textbf{LLM 5}} & \multicolumn{2}{c}{\textbf{LLM 6}} \\ \midrule
 & \textit{ZS} & \textit{FS} & \textit{ZS} & \textit{FS} & \textit{ZS} & \textit{FS} & \textit{ZS} & \textit{FS} & \textit{ZS} & \textit{FS} & \textit{ZS} & \textit{FS} \\ \midrule
\textbf{P7}  & 0.00 & 12.32 & 11.09 & 29.09 & 0.00 & 18.51 & 2.38 & 11.36 & 24.62 & 21.26 & 18.59 & 23.55 \\
\textbf{P9}  & 0.00 & 10.68 & 21.53 & 31.06 & 0.00 & 19.62 & 2.82 & 11.33 & 24.19 & 22.46 & 19.14 & 22.52 \\ 
\midrule
\textbf{MaxP} & 0.00 & 12.32 & 21.53 & \textbf{31.06} & 0.00 & 19.62 & 2.82 & 11.36 & 24.62 & 22.46 & 19.14 & 23.55 \\
\textbf{AvgP} & 0.00 & 11.50 & 16.31 & \textbf{30.08} & 0.00 & 19.07 & 2.60 & 11.35 & 24.41 & 21.86 & 18.87 & 23.04 \\
\textbf{CPS} & 0.00 & 12.22 & 20.41 & \textbf{30.75} & 0.00 & 19.51 & 2.81 & 11.36 & 24.57 & 22.33 & 19.09 & 23.43 \\
\bottomrule
\end{tabular}
\caption{LLM Results with Zero-shot (ZS) and Few-shot (FS) Evaluations for the Lexical Substitution Task.}
\label{tab:ls_results}
\end{table}

\begin{table}[ht]
\centering
\setlength{\tabcolsep}{4pt}  
\begin{tabular}{c|c c|c c|c c|c c|c c|c c}
\toprule
 & \multicolumn{2}{c|}{\textbf{LLM 1}} & \multicolumn{2}{c|}{\textbf{LLM 2}} & \multicolumn{2}{c|}{\textbf{LLM 3}} & \multicolumn{2}{c|}{\textbf{LLM 4}} & \multicolumn{2}{c|}{\textbf{LLM 5}} & \multicolumn{2}{c}{\textbf{LLM 6}} \\ \midrule
 & \textit{ZS} & \textit{FS} & \textit{ZS} & \textit{FS} & \textit{ZS} & \textit{FS} & \textit{ZS} & \textit{FS} & \textit{ZS} & \textit{FS} & \textit{ZS} & \textit{FS} \\ \midrule
\textbf{P7}  & 0.00 & N/A & 22.31 & 19.99 & 24.66 & 20.46 & 26.81 & 23.51 & 7.98 & 30.45 & 21.84 & 28.32 \\
\textbf{P9}  & 0.00 & N/A & 22.19 & 20.57 & 27.01 & 20.73 & 27.19 & 24.49 & 9.83 & 26.63 & 22.42 & 29.36 \\
\midrule
\textbf{MaxP} & 0.00 & 0.00 & 22.31 & 20.57 & 27.01 & 20.73 & 27.19 & 24.49 & 9.83 & \textbf{30.45} & 22.42 & 29.36 \\
\textbf{AvgP} & 0.00 & N/A & 22.25 & 20.28 & 25.84 & 20.60 & 27.00 & 24.00 & 8.91 & 28.54 & 22.13 & \textbf{28.84} \\
\textbf{CPS} & 0.00 & N/A & 22.30 & 20.51 & 26.69 & 20.70 & 27.14 & 24.37 & 9.74 & \textbf{29.87} & 22.35 & 29.21 \\
\bottomrule
\end{tabular}
\caption{LLM Results with Zero-shot (ZS) and Few-shot (FS) Evaluations for the Summarization Task.}
\label{tab:su_results}
\end{table}

\begin{table}[ht]
\centering
\setlength{\tabcolsep}{4pt}  
\begin{tabular}{c|c c|c c|c c|c c|c c|c c}
\toprule
 & \multicolumn{2}{c|}{\textbf{LLM 1}} & \multicolumn{2}{c|}{\textbf{LLM 2}} & \multicolumn{2}{c|}{\textbf{LLM 3}} & \multicolumn{2}{c|}{\textbf{LLM 4}} & \multicolumn{2}{c|}{\textbf{LLM 5}} & \multicolumn{2}{c}{\textbf{LLM 6}} \\ \midrule
 & \textit{ZS} & \textit{FS} & \textit{ZS} & \textit{FS} & \textit{ZS} & \textit{FS} & \textit{ZS} & \textit{FS} & \textit{ZS} & \textit{FS} & \textit{ZS} & \textit{FS} \\ \midrule
\textbf{P8}  & 20.61 & 53.51 & 15.51 & 68.82 & 23.98 & 55.35 & 39.49 & 66.90 & 40.46 & 38.21 & 48.31 & 31.43 \\
\textbf{P10}  & 7.45  & 56.30 & 13.12 & 80.27 & 28.51 & 55.35 & 40.07 & 67.82 & 34.91 & 32.72 & 28.20 & 33.20 \\
\midrule
\textbf{MaxP} & 20.61 & 56.30 & 15.51 & \textbf{80.27} & 28.51 & 55.35 & 40.07 & 67.82 & 40.46 & 38.21 & 48.31 & 33.20 \\
\textbf{AvgP} & 14.03 & 54.91 & 14.32 & \textbf{74.55} & 26.25 & 55.35 & 39.78 & 67.36 & 37.69 & 35.47 & 38.26 & 32.32 \\
\textbf{CPS} & 19.25 & 55.51 & 15.32 & \textbf{75.67} & 27.86 & 55.35 & 39.95 & 67.51 & 39.34 & 37.16 & 43.45 & 32.91 \\
\bottomrule
\end{tabular}
\caption{LLM Results with Zero-shot (ZS) and Few-shot (FS) Evaluations for the Named Entity Recognition Task.}
\label{tab:ner_results}
\end{table}

\begin{table}[ht]
\centering
\setlength{\tabcolsep}{4pt}  
\begin{tabular}{c|c c|c c|c c|c c|c c|c c}
\toprule
 & \multicolumn{2}{c|}{\textbf{LLM 1}} & \multicolumn{2}{c|}{\textbf{LLM 2}} & \multicolumn{2}{c|}{\textbf{LLM 3}} & \multicolumn{2}{c|}{\textbf{LLM 4}} & \multicolumn{2}{c|}{\textbf{LLM 5}} & \multicolumn{2}{c}{\textbf{LLM 6}} \\ \midrule
 & \textit{ZS} & \textit{FS} & \textit{ZS} & \textit{FS} & \textit{ZS} & \textit{FS} & \textit{ZS} & \textit{FS} & \textit{ZS} & \textit{FS} & \textit{ZS} & \textit{FS} \\ \midrule
\textbf{P8}  & 0.00 & 25.24 & 14.52 & 35.82 & 6.66 & 28.88 & 7.14 & 35.88 & 30.15 & 26.04 & 14.25 & 43.50 \\
\textbf{P10}  & 0.00 & 23.30 & 15.55 & 35.61 & 8.40 & 28.54 & 11.55 & 35.90 & 28.83 & 35.25 & 19.92 & 40.43 \\
\midrule
\textbf{MaxP} & 0.00 & 25.24 & 15.55 & 35.82 & 8.40 & 28.88 & 11.55 & 35.90 & 30.15 & 35.25 & 19.92 & \textbf{43.50} \\
\textbf{AvgP} & 0.00 & 24.27 & 15.04 & 35.72 & 7.53 & 28.71 & 9.35 & 35.89 & 29.49 & 30.65 & 17.09 & \textbf{41.97} \\
\textbf{CPS} & 0.00 & 25.00 & 15.47 & 35.78 & 8.33 & 28.83 & 11.30 & 35.90 & 29.95 & 33.63 & 19.36 & \textbf{42.83} \\
\bottomrule
\end{tabular}
\caption{LLM Results with Zero-shot (ZS) and Few-shot (FS) Evaluations for the Relation Extraction Task.}
\label{tab:rel_results}
\end{table}




\end{appendices}


\clearpage
\bibliography{sn-bibliography}

\end{document}